\documentclass[sn-vancouver,Numbered,iicol]{sn-jnl}% Vancouver Reference Style
\usepackage{graphicx}%
\usepackage{multirow}%
\usepackage{amsmath,amssymb,amsfonts}%
\usepackage{amsthm}%
\usepackage{mathrsfs}%
\usepackage[title]{appendix}%
\usepackage{xcolor}%
\usepackage{textcomp}%
\usepackage{manyfoot}%
\usepackage{booktabs}%
\usepackage{algorithm}%
\usepackage{algorithmicx}%
\usepackage{algpseudocode}%
\usepackage{listings}%

\usepackage{colortbl}

\usepackage{graphicx}
\usepackage{subcaption}

\usepackage{makecell}
\usepackage{mathrsfs}

\usepackage{lineno}
\theoremstyle{thmstyleone}%
\theoremstyle{thmstyletwo}%

\theoremstyle{thmstylethree}%

\begin{document}
	
	% \title[Article Title]{Article Title}
	\title[Article Title]{Improving 3D Finger Traits Recognition via Generalizable Neural Rendering}
	

	\author[1]{\fnm{Hongbin} \sur{Xu}}
	
	\author[1]{\fnm{Junduan} \sur{Huang}}
	
	\author[1]{\fnm{Yuer} \sur{Ma}}
	
	\author[1]{\fnm{Zifeng} \sur{Li}}
	
	\author*[1]{\fnm{Wenxiong} \sur{Kang}}\email{auwxkang@scut.edu.cn}
	
	\affil*[1]{\orgdiv{School of Automation Science and Engineering}, \orgname{South China University of Technology}, \orgaddress{Wushan Road}, \city{Guangzhou}, \postcode{510641}, \state{Guangdong}}
	
	
	\abstract{3D biometric techniques on finger traits have become a new trend and have demonstrated a powerful ability for recognition and anti-counterfeiting. 
		Existing methods follow an explicit 3D pipeline that reconstructs the models first and then extracts features from 3D models.
		However, these explicit 3D methods suffer from the following problems: 1) Inevitable information dropping during 3D reconstruction; 2) Tight coupling between specific hardware and algorithm for 3D reconstruction.
		It leads us to a question: Is it indispensable to reconstruct 3D information explicitly in recognition  tasks?
		Hence, we consider this problem in an implicit manner, leaving the nerve-wracking 3D reconstruction problem for learnable neural networks with the help of neural radiance fields (NeRFs).
		We propose FingerNeRF, a novel generalizable NeRF for 3D finger biometrics.
		To handle the shape-radiance ambiguity problem that may result in incorrect 3D geometry, we aim to involve extra geometric priors based on the correspondence of binary finger traits like fingerprints or finger veins.
		First, we propose a novel Trait Guided Transformer (TGT) module to enhance the feature correspondence with the guidance of finger traits.
		Second, we involve extra geometric constraints on the volume rendering loss with the proposed Depth Distillation Loss and Trait Guided Rendering Loss.
		To evaluate the performance of the proposed method on different modalities, we collect two new datasets: SCUT-Finger-3D with finger images and SCUT-FingerVein-3D with finger vein images.
		Moreover, we also utilize the UNSW-3D dataset with fingerprint images for evaluation.
		In experiments, our FingerNeRF can achieve 4.37\% EER on SCUT-Finger-3D dataset, 8.12\% EER on SCUT-FingerVein-3D dataset, and 2.90\% EER on UNSW-3D dataset, showing the superiority of the proposed implicit method in 3D finger biometrics.
        For access to our project page and code, please visit our \href{https://scut-bip-lab.github.io/fingernerf/}{project page}.
	}
	
	
	\maketitle

	% \pagewiselinenumbers% 按页重新编号 
	% \switchlinenumbers
	
	\section{Introduction}
	\label{introduction}
	
	With the development of biometrics, many advanced biometrics methods are proposed. Among them, three-dimention (3D) biometrics is one of the most potential mainstreams due to the following advantages: 1) 3D biometric traits have more identity-discriminating information, which is directly related to authentication accuracy; 2) 3D biometric traits are more robust than two-dimention (2D) biometric traits; because its more comprehensive information can avoid the impacts caused by the capture-perspective variation; 3) 3D biometric traits are more difficult to forge, therefore, have better anti-spoofing capabilities.
	
	Among the 3D biometrics, 3D finger biometrics is attracting more and more attention from the biometrics community, either academic field or industrial field. This is because there are some common and convenient used biometric trait on the finger, which have been studied for a long time and now they are in a bottleneck period, running into some limitations. Therefore, the biometrics methods based on their 3D forms that can overcome these limitations and have better performance are becoming the naturally promising research orientation. The main 3D finger traits are 3D fingerprint \cite{Cui2022Monocular3F,Lin2018TetrahedronBF}, 3D finger knuckle \cite{Cheng2021DeepFC,Cheng2020ContactlessBI} and 3D finger vein \cite{Kang2020StudyOA,Zhan20203DFV,Xu2022EndowingRI}.
	
	Since these 3D finger biometric traits are all located on the finger and can be easily captured simultaneously, some researchers treat them as a whole, namely 3D finger biometrics \cite{Yang2022ANS,Yang2021LFMB3DFBAL}. The advantages of 3D finger biometrics are:
	
	\begin{enumerate}
		\item \textbf{Comprehensive}: 3D finger biometrics includes multiple finger biometric traits, hence this manner has more comprehensive information than the single traits. The more comprehensive information leads to higher accuracy. 
		\item \textbf{Robust}: Since 3D finger biometrics includes multiple finger biometric traits, its authentication system will be more robust than using the single finger trait, especially when some of the certain finger traits are contaminated. For example, the sweaty fingerprint will affect the biometrics system’s performance, but this case can be well tackled by 3D finger biometrics since the fingerprint is not the only source of all distinguishing information.
		\item \textbf{Anti-counterfeiting}: 3D finger biometrics make it harder for counterfeiting, because the forgers need to obtain more comprehensive information to forge 3D finger. This will further improve the security of biometrics system.
	\end{enumerate}

	In this work, \textbf{we aim to handle the recognition problem of 3D finger biometrics}.
	The standard pipeline of existing 3D methods mainly includes the following procedures: 
	1) \emph{3D Reconstruction:} Recovering the 3D information in an active manner \cite{kumar2013towards} or passive manner \cite{yang2021lfmb, Dong2023SynthesisOM}. 
	The active methods adopt specially designed 3D imaging devices to obtain the 3D information of the finger, such as photometric stereo on fingerprints \cite{kumar2013towards} and finger knuckles \cite{Cheng2020ContactlessBI}.
	The passive methods attempt to recover the 3D information from given the single-view image \cite{Cui2022Monocular3F} or multi-view images \cite{Kang2020StudyOA, yang2021lfmb, Xu2022EndowingRI}.
	2) \emph{Feature Extraction:} Extracting features from the reconstructed 3D models \cite{yang2021lfmb,lin2018contactless}.
	Many superior methods are proposed considering the reconstructed modality of 3D finger traits, such as multi-view convolutional networks for handling multi-view images \cite{yang2021lfmb,lin2018contactless}, 3D point cloud perception networks for reconstructed point clouds \cite{Xu2022EndowingRI}, and etc.
	Thanks to these novel efforts, great progresses have been witnessed in the research field of 3D finger biometrics.
	
	\begin{figure*}[t]%
		\centering
		\includegraphics[width=0.9\textwidth]{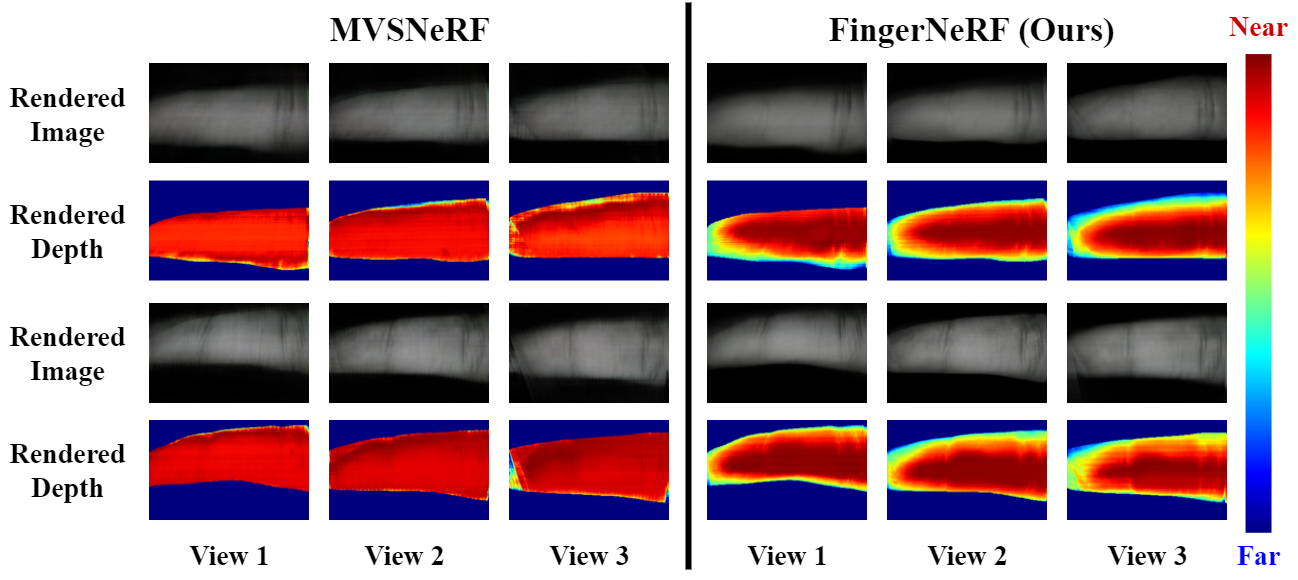}
		\caption{Illustration of the shape-radiance ambiguity problem in existing generalizable NeRFs, like MVSNeRF \cite{chen2021mvsnerf}.}
		\label{fig:motivation}
		\vspace{-0.4cm}
	\end{figure*}

	However, the existing methods are mostly based on explicit 3D reconstruction techniques and suffer from following limitations:
	\begin{enumerate}
		\item \textbf{Information Dropping during 3D reconstruction}: The tedious process of explicit 3D reconstruction inevitably drops some information, and the final performance in recognition task is highly dependent on the accuracy and completeness of the reconstructed 3D models.
		\item \textbf{Tight coupling between hardware and algorithm}: The tight coupling between imaging device and explicit 3D reconstruction algorithm makes the existing methods can only be used on certain modality and customized devices, unable to be migrated to other modalities or devices.
	\end{enumerate}

	Considering these problems caused by explicit 3D reconstruction, we wonder: \textbf{Is it indispensable to reconstruct the 3D finger model explicitly?}
	Aiming at serving for 3D finger biometrics, the excessive concerns on 3D finger reconstruction may not pay handsome dividends because we only need the identity-discriminating information that can serve the 3D finger trait recognition.
	Consequently, we would like to consider this problem differently in an \textbf{implicit} way.
	Instead of concentrating on modeling the 3D space explicitly, \textbf{we can alternatively model the 3D space as a neural radiance field (NeRF) \cite{mildenhall2021nerf} via learnable parameters implicitly}.
	NeRF and its following works \cite{mildenhall2021nerf, liu2020neural, martin2021nerf} have shown impressive performance on novel view synthesis of a 3D scene by implicitly encoding the color and volume density on rays through neural networks.
	The implicit NeRF-based representation can release us from the tedious procedure of building a explicit 3D reconstruction pipeline.
	In this way, the 3D scene is implicitly recovered by training the NeRF representation with a proxy task of neural rendering.

	Though the NeRF representation sounds feasible, there is another problem: the original representation of NeRF requires abundant number of surrounding viewpoints and needs per-scene optimization.
	It is in conflict with the open-set setting in finger trait recognition task, that usually requires generalizable ability towards different scenes.
	Furthermore, the few-shot learning in 3D finger biometrics further limits the input number of views to a sparse setting.
	Consequently, we aim to design a NeRF-based representation that is \textbf{not only generalizable towards unseen subjects of 3D finger biometrics but also able to handle input images of sparse viewpoints}.
	
	Recent advances in generalizable neural rendering \cite{chen2021mvsnerf, yu2021pixelnerf, wang2021ibrnet}, borrows the cost volume representation used in multi-view stereo (MVS) \cite{yao2018mvsnet,yao2019recurrent,xu2021self,xu2021digging} as the input of conditional NeRF.
	Since the cost volume is constructed by warping the 2D image features of nearby views (infered by CNNs) on sweeping depth planes in the reference view's frustum, the matching relationship on epipolar lines is inherently modeled via homography function.
	Unlike MVS that uses cost volume to infer depth maps, MVSNeRF \cite{chen2021mvsnerf}
	uses the cost volume to encode the per-scene geometry and appearance via neural encoding volume.
	However, the \textbf{shape-radiance ambiguity problem} occurs when these generalizable NeRFs \cite{chen2021mvsnerf} are directly trained by rendering finger trait images, as shown in Fig. \ref{fig:motivation}.
	Although the exisiting generalizable NeRF can render reasonable finger images, the rendered depth map seems to be incorrect in 2-nd/4-th row of the figure, due to the overfitting effect of radiance.
	The shape-radiance ambiguity problem can be resorted to 2 reasons: 1) \emph{The images of finger biometrics have less abundant textures compared with natural images, making it difficult to excavate the cross-view correspondence.} 2) \emph{The training pipeline of previous methods are merely based on differentiable neural rendering on multi-view images, lacking extensive constraints on the geometric property of 3D shapes.}

	To handle the aforementioned shape-radiance ambiguity problem, we aim to involve extra 3D shape priors based on the characteristics of 3D finger biometrics.
	The feasibility of finger biometrics for recognizing certain person is based on the assumption that the finger trait (e.g. fingerprint/finger vein) is inherently a simplified distinguishable representation compared with raw images.
	Consequently, we can involve the correspondence of these finger traits among views as constraints to regularize the training of NeRF representation.
	First, we propose Trait Guided Transformer (TGT) to enhance the cross-view feature maps via self-attention guided by finger traits.
	It can involve the correspondence of finger traits to guide the construction of cost volume.
	Second, we involve extra depth constraints by incorporating the neural rendering training with the proposed Depth Distillation Loss and Trait Guided Rendering Loss.
	DD-Loss can distill the coarse geometric prior estimated by large model in monocular depth estimation \cite{ranftl2020towards} to the rendered depth maps irrespective of the scale difference.
	TG-Loss can regularize the neural rendering photometric loss with finger trait clues, implicitly regularizing the correspondence of finger traits across views on the ray.
	
	Concretely, our contributions are summarized as follows:
	\begin{itemize}
		\item We firstly handle the problem of 3D finger biometrics via an implicit NeRF-based representation.
		It can implicitly model the 3D information via differentiable rendering.
		
		\item We propose FingerNeRF, a novel generalizable NeRF-based method for 3D finger biometrics, that has following merits: a) Generalizing towards unseen subjects following the open-set setting in biometrics; b) Requiring only sparse inputs of viewpoints (3 in default); c) Remedying the shape-radiance ambiguity and render reasonable depth maps.
		
		\item  To handle shape-radiance ambiguity, we involve extra geometric constraints in FingerNeRF by two techniques: a) We propose Trait Guided Transformer (TGT) module to enhance the cross-view feature maps via traits on epipolar line; b) We insert the constraints on depth maps via the proposed DD-Loss and TG-Loss.
		
		\item To evaluate the effectiveness of FingerNeRF, we collect 2 datasets with different modalities: SCUT-Finger-3D with finger images and SCUT-FingerVein-3D with finger vein images.
		Furthermore we also evaluate the FingerNeRF on existing 3D fingerprint dataset UNSW-3D.
		The experimental results on these datasets support the superiority of the proposed methods.
	\end{itemize}
	
		The rest of this paper is organized as follows.
		Section \ref{related_work} outlines the related work of 3D Verification on finger biometrics and neural rendering.
		Section \ref{methods} firstly presents the preliminary of volume rendering techniques, and then provide detailed description of the proposed FingerNeRF framework.
		In Section \ref{experiment}, we firstly introduce the three utilized datasets and the evaluation benchmarks, and then discuss the experimental results on these datasets comparing with state-of-the-art methods on 3D recognition.
		Section \ref{limitation} further discusses the limitation of the proposed FingerNeRF and Section \ref{conclusion} concludes the paper.

	\section{Related Work}
	\label{related_work}
	
	\subsection{3D Verification on Finger Biometrics}

	For evaluating the authentication performance of 3D fingerprint and the compatibility of 2D and 3D fingerprint, Zhou \emph{et al.} \cite{Zhou2014PerformanceEO} establish a dataset that includes both 3D fingerprint and its corresponding 2D fingerprint. This is one of the earliest 3D fingerprint biometrics exploration and the results show that the performance of 3D fingerprint authentication is comparable to that of the traditional 2D fingerprint.
	Cui \emph{et al.} \cite{Cui2022Monocular3F} propose an approach for 3D finger reconstruction and unwarping method. First, they sent the preprocessed fingerprint into the network to estimate its surface gradients; then these estimated gradients are used for 3D shape reconstruction, and finally the fingerprint is unwrapped. Experimental results show that the proposed unwarping method can reduce perspective distortion, which is significant for fingerprint matching.
	Recently, Dong \emph{et al.} \cite{Dong2023SynthesisOM} propose a method for accurately synthesizing the multi-view 3D fingerprint to develop a large-scale multi-view fingerprint dataset, which can ensure a high degree of freedom and realness of synthetic 3D fingerprint model which balance the computation time. This is the first attempt to synthesize 3D fingerprint and will also the explore for unlocking a range of new possibilities and new research directions about 3D fingerprint.

	In addition to 3D fingerprint, 3D finger knuckle is also one of the main 3D finger biometric traits, Cheng \emph{et al.} \cite{Cheng2020ContactlessBI} propose the first 3D finger knuckle dataset for public scientific research, basing on photometric stereo approach. In their work, a new feature descriptor for extracting discriminative 3D finger knuckle is also proposed. 
	For addressing the challenges of 3D finger knuckle feature extraction by deep network, Cheng \emph{et al.} \cite{Cheng2021DeepFC} further propose a FKNet, which is demonstrated by the experimental results that superior than the SOTA handcraft finger knuckle feature. Recently, there is a latest follow work \cite{Cheng2023Advancing3F}, which can achieve outperforming results in classification and identification tasks under the practical feature comparison scenario.

	Different from the above 3D finger biometric traits that locate on the finger skin, Kang \emph{et al.} \cite{Kang2020StudyOA} explore the 3D finger vein biometrics, which is underneath the finger skin. In this work, 3D finger vein is reconstructed basing on the prior that the finger’s cross section is an approximate ellipse. Then the, for using the excellent feature extraction ability of CNN, the reconstructed 3D finger vein is unfolded into a finger vein texture image and a finger shape texture image. Finally, deep CNN is used to extract the feature of both finger texture and finger shape images and achieve satisfy results. 
	For further solving the finger posture variation problem, Xu \emph{et al.} \cite{Xu2022EndowingRI} propose a silhouette-based 3D finger vein reconstruction method, namely 3D rotation invariant network (3DFVSNet). To evaluate the rotation equivariance of rotation group and rotation invariance of the features extracted by 3DFVNet, the visualization experiment is conducted and the experimental results demonstrate their effeteness.

	Not limited by a certain 3D finger biometric trait, Yang \emph{et al.} \cite{Yang2021LFMB3DFBAL,Yang2022ANS} propose an multi-view and multi-biometric-traits capture device. Based on the captured multi-view finger images, the 3D finger mesh model is reconstructed and then used for authentication. This work breaks the boundaries of 3D finger biometric traits, providing a new way for 3D finger biometrics.
	
	\subsection{Neural Rendering}
	
	Recent progresses in neural scene representations have been proposed to realize novel view synthesis and geometric reconstruction \cite{lombardi2019neural,bi2020deep,mildenhall2021nerf}. 
	The pioneering work of NeRF \cite{mildenhall2021nerf} firstly utilizes MLPs to represent the radiance field and optimize these MLPs via differentiable neural rendering on multi-view images.
	The implicit representation of NeRF can model the 3D scenarios and render photo-realistic images from arbitrary viewpoints.
	However, the requirement of dense multi-view inputs and per-scene optimization make the NeRF representation not suitable for realistic applications, which usually provides only sparse viewpoints and need to generalize towards unseen scenes without finetuning, like the 3D finger biometrics of this paper.
	As a result, some researches \cite{yu2021pixelnerf,wang2021ibrnet,chen2021mvsnerf} aiming at endowing the NeRFs with generalization ability have been proposed.
	PixelNeRF \cite{yu2021pixelnerf} conditions a NeRF representation on image features extracted from a fully-connected convolutional neural network.
	IBRNet \cite{wang2021ibrnet} combines the techniques of image-based rendering and NeRF, and aggregate information from source views along a given ray to compute the radiance filed.
	MVSNeRF \cite{chen2021mvsnerf} is a combination of Multi-view Stereo (MVS) networks \cite{yao2018mvsnet,yao2019recurrent,xu2021self,xu2021digging} and NeRF.
	MVS is a classical computer vision problem, aiming to achieve dense 3D reconstruction of scenarios using multi-view images.
	Unlike MVS that computes a plane-swept cost volume at the reference view for depth estimation, MVSNeRF leverages the feature extracted from the cost volume to condition the NeRF and achieve superior generalization ability in neural rendering.
	Despite the impressing progress in neural rendering techniques, the shape-density ambiguity problem still restricts the application of neural rendering techniques, especially the 3D finger biometrics in this paper, as discussed in Section \ref{introduction} and Fig. \ref{fig:motivation}.
	Our approach aims at handling this shape-density ambiguity problem via inserting extra prior of finger traits in this paper and meantime achieve the generalization ability given sparse views on unseen scenarios.
	
	\section{Method}
	\label{methods}
	
\begin{figure*}[t]%
		\centering
		\includegraphics[width=\textwidth]{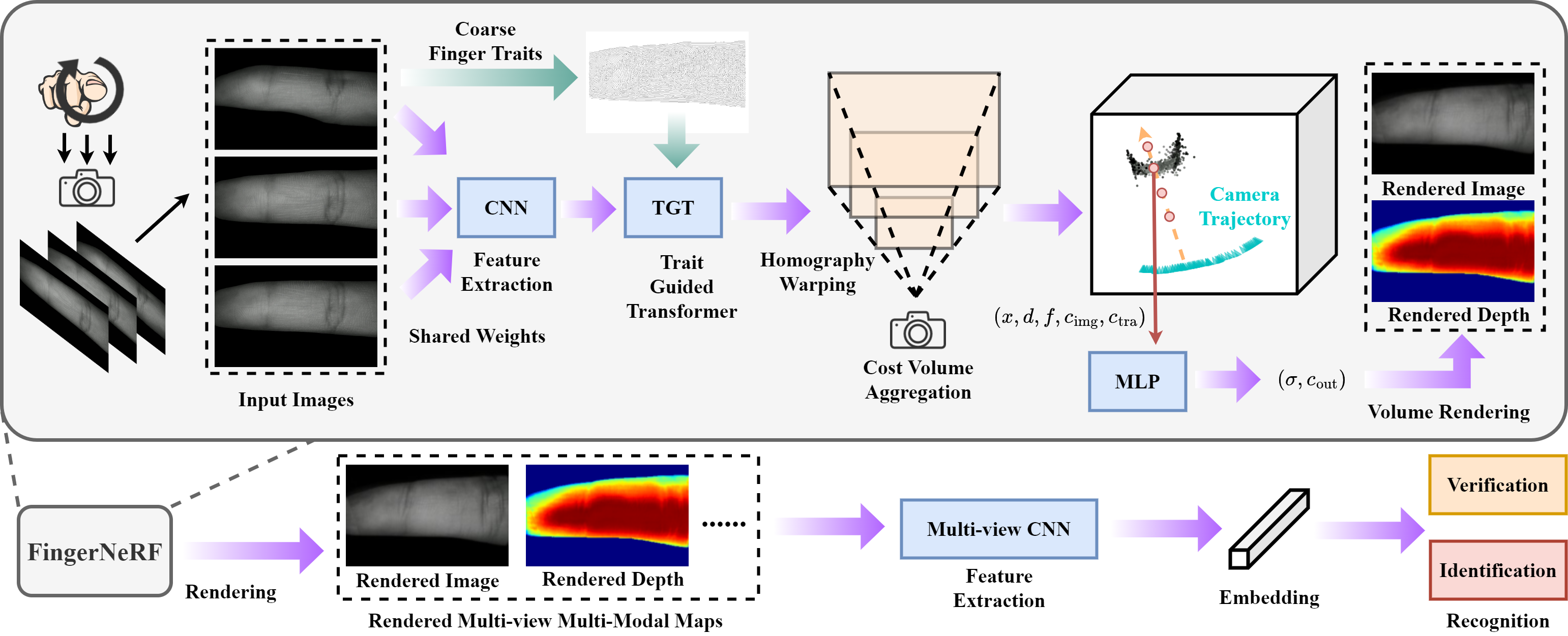}
		\caption{Brief overview of the proposed FingerNeRF.}
		\label{fig:overview}
		\vspace{-0.4cm}
	\end{figure*}

	In this section, we introduce the proposed FingerNeRF.
	A brief overview of FingerNeRF is presented in Fig. \ref{fig:overview}.
	We first revisit the preliminary knowledge about Neural Radiance Fields (NeRFs) in Section \ref{methods:preliminary}.
	Then, we discuss the limitation of existing generalizable NeRFs and introduce the details of the proposed FingerNeRF in Section \ref{methods:fingernerf}.
	
	\subsection{Preliminary}
	\label{methods:preliminary}
	
	\subsubsection{NeRF and Volume Rendering}
	We first briefly review the NeRF representation \cite{mildenhall2021nerf}.
	A scene is encoded as a continuous volumetric radiance field $\phi$ of color and density in NeRF.
	Given the input of a 3D point on the ray ${x} \in \mathbb{R}^3$ and view direction vector ${d} \in \mathbb{R}^3$, the NeRF $\phi$ returns the volume density $\sigma \in \mathbb{R}$ and color ${c} \in \mathbb{R}^3$: $\phi({x}, {d}) = (\sigma, {c})$.
	
	By sampling 3D points of the rays on 2D pixels and calculating the view direction from pixel coordinate and camera center, we can build a volumetric radiance field, that can then be rendered into a 2D image via:
	\begin{equation}
		\hat{{C}} ({r}) = \int_{t_n}^{t_f} \text{exp} (- \int_{t_n}^t \sigma(s) ds) \sigma(t) {c}(t)
		\label{eq1}
	\end{equation}
	where the occlusion along the ray is handled via: $\text{exp} (- \int_{t_n}^t \sigma(s) ds)$.
	A camera ray can be parameterized as ${r} (t) = {o} + t {d}$, where ${o} \in \mathbb{R}^3$ is the ray origin (camera center) and ${d} \in \mathbb{R}^3$ is the ray direction vector computed from the pixel coordinate and camera center.
	Assume that the scene is bounded by the nearest and farthest depth values: $[ t_n, t_f ]$, the integral is computed along the ray ${r}$ between these depth boundaries.
	In practice, the integral in the equation is approximated with numerical quadrature by sampling points discretely along the ray ${r}$.
	
	\subsubsection{Training Loss}
	With the help of the volume rendering function, the rendered pixel value on camera ray ${r}$ can be compared against corresponding ground truth pixel ${C} ({r})$ in original images.
	The photometric loss to supervise the NeRF rendering results is computed as follows:
	\begin{equation}
		{L} = \sum_{{r} \in {R}({P})} \| \hat{{C}} ({r}) - {C} ({r})\|_2^2
		\label{eq2}
	\end{equation}
	where ${R}({P})$ is the set of all camera rays in the 2D image on target pose ${P}$.
	
	\subsubsection{MVSNeRF}
	To endow the generalization ability of MVSNet \cite{yao2018mvsnet} into the NeRF representation, the cost volume is used to condition the volume rendering process in MVSNeRF.
	The multi-view feature maps are extracted via a deep 2D CNN.
	Given the input image $I_i \in \mathbb{R}^{N_H \times N_W \times 3}$ on view $i$, the output feature map of the 2D CNN is $F_i \in \mathbb{R}^{N_H/4 \times N_W/4 \times N_C}$.
	$N_H$ and $N_W$ are the height and width. $N_C$ is the output feature channels.
	Given the camera intrinsic $K$ and extrinsic parameters $[R, t]$, the homography function can be computed: 
	\begin{equation}
		\mathcal{H}_i (z) = K_i \cdot (R_i \cdot R_1^T + \frac{(t_1 - t_i) \cdot n_1^T}{z}) \cdot K_1^{-1}
		\label{eq3}
	\end{equation}
	where $\mathcal{H}_i(z)$ is the matrix warping from view $i$ to the reference view ($i=1$) at depth $z$ and normal $n_1$.
	Then the feature maps can be warped to the reference view by: $G_{i,z}(u,v) = F_i (\mathcal{H}_i(z)[u,v,1]^T)$.
	$G_{i,z}$ is the warped feature map at depth $z$, and $(u,v)$ means the pixel coordinates.
	By stacking a series of depth planes $z$, the cost volume can be constructed as:
	\begin{equation}
		P(u,v,z) = \text{var} (G_{i,z}(u,v))
		\label{eq4}
	\end{equation}
	where $\text{var}$ computes the variance across different views.
	Afterwards, the cost volume will be fed to a 3D U-Net B which can effectively aggregate geometry features encoded in the cost volume, leading to a meaningful neural encoding volume $S$: $S = B(P)$.
	
	The final representation of MVSNeRF is formulated as: $\sigma, c = A(x, d, S(x), c)$.
	$S(x)$ is trilinearly interpolated from the neural encoding volume $S$ at the location of 3D point $x$.
	$(u_i, v_i)$ means the pixel location when projecting 3D point $x$ to view $i$.
	$c$ concatenates the color of original image $I(u_i, v_i)$ on all views.
	$A$ is a series of MLP layers.
	Then the MVSNeRF can be trained with photometric loss in Eq. \ref{eq2}.
	
	\subsection{FingerNeRF}
	\label{methods:fingernerf}
	
	As Section \ref{introduction} and Fig. \ref{fig:motivation} show, the shape-radiance ambiguity problem in MVSNeRF may return incorrect 3D geometry and overfit on the radiance rendering process.
	Before introducing the solutions, we can first \textbf{rethink the architecture of generalizable NeRFs (e.g. MVSNeRF) as an ``\emph{encoder-decoder}" structure}: 
	\begin{enumerate}
		\item The encoder is a 3D reconstruction network, such as MVSNet used in MVSNeRF.
		The intermediate implicit representation of cost volume is preserved as output rather the estimated depth maps.
		
		\item The decoder is a NeRF conditioned on the feature of cost volume trilinearly interpolated on the sampled 3D points.
	\end{enumerate}
	
	Specifically, the shape-radiance ambiguity problem can then be traced back to the encoder and decoder respectively:
	
	\begin{enumerate}
		\item In the encoder, the 3D reconstruction network can not recover the cross-view correspondence effectively due to the particularity of finger trait images. Consequently, we aim to involve the correspondence prior of finger traits which inherently preserves personal identity into the construction of cost volume. (Section \ref{methods:fingernerf:trait_transformer})
		
		\item In the decoder, the rendering process is only supervised by the RGB images.
		Without direct constraints on 3D geometry, the weighted coupling between volume density and color suffers from imbalance caused by the overfitting effect on color images.
		Consequently, we can design a series of geometric constraints as a regularization to the neural rendering. (Section \ref{methods:fingernerf:extra_priors})
		
	\end{enumerate}
	
	\subsubsection{Trait Guided Transformer}
	\label{methods:fingernerf:trait_transformer}
	
	\begin{figure*}[t]%
		\centering
		\includegraphics[width=\textwidth]{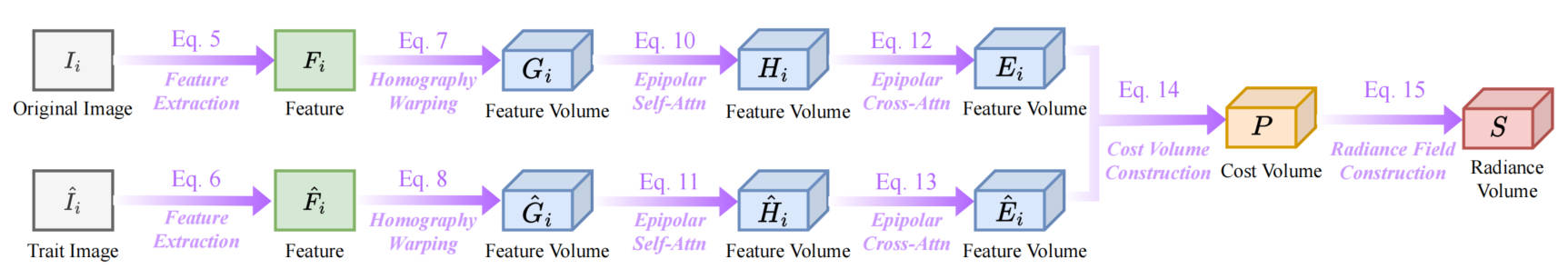}
		\caption{Detailed architecture of the network modules in our proposed FingerNeRF.}
		\label{fig:fingernerf_formula}
		\vspace{-0.4cm}
	\end{figure*}
	
	In this section, we introduce the Trait Guided Transformer (TGT) which aims to involve extra correspondence prior of finger traits into the cost volume.
	The details of the network modules and the related formulas are shown in Figure \ref{fig:fingernerf_formula}.
	Denote that the input multi-view images are $I_i \in \mathbb{R}^{N_H \times N_W \times 3}$, where $i$ is the index of views.
	From the multi-view images of finger, we can extract the finger trait images $\hat{I}_i \in \mathbb{R}^{N_H \times N_W \times 1}$ from the original images.
	For simplicity, the finger trait $\hat{I}_i$ is a binary image representing whether the trait exists or not.
	The finger trait is extracted via traditional methods in previous works, i.e. fingerprint \cite{kizrak2011new}, finger vein \cite{miura2004feature}.
	
	\noindent\textbf{Feature Extraction}:
	We respectively use two different CNNs to extract 2D feature maps from original finger images and trait images:
	\begin{equation}
		F_i = \psi (I_i)
		\label{eq5}
	\end{equation}
	\begin{equation}
		\hat{F}_i = \hat{\psi}(\hat{I}_i)
		\label{eq6}
	\end{equation}
	where the network $\psi$ extracts 2D feature map $F_i \in \mathbb{R}^{N_H \times N_W \times N_{C1}}$ from original image $I_i$, and the network $\hat{\psi}$ extracts 2D feature map $\hat{F}_i \in \mathbb{R}^{N_H \times N_W \times N_{C1}}$ from trait image $\hat{I}_i$.
	$N_{C1}$ is the number of feature channels.
	
	\noindent\textbf{Feature Volume Construction}:
	Utilizing the homography warping function in Eq. \ref{eq3}, we can then build the warped feature volume on the reference view:
	\begin{equation}
		G_{i,z} (u,v) = F_i (\mathcal{H}_i(z)[u,v,1]^T)
		\label{eq7}
	\end{equation}
	\begin{equation}
		\hat{G}_{i,z} (u,v) = \hat{F}_i (\mathcal{H}_i(z)[u,v,1]^T) 
		\label{eq8}
	\end{equation}
	where $\mathcal{H}_i(z)$ is the homography mapping matrix on the depth plane $z$.
	$G_{i,z}$ and $\hat{G}_{i,z}$ are the warped feature matrix on the depth plane $z$.
	Then the feature volume ${G}_i \in \mathbb{R}^{N_H \times N_W \times N_D \times N_{C1}}$ and $\hat{{G}}_i  \in \mathbb{R}^{N_H \times N_W \times N_D \times N_{C1}}$ of view $i$ can be built by traversing depth planes ranging from the nearest depth value to the farthest one.
	$N_D$ is the number of depth hypotheses.
	Instead of directly computing the variance of ${G}_i$ like MVSNeRF in Eq. \ref{eq4}, we utilize the finger trait feature volume $\hat{{G}}_i$ to guide the cost volume construction.
	
	\noindent\textbf{Epipolar Self-attention}:
	Given the scaled product attention as follows:
	\begin{equation}
		\text{Att} ({Q}, {K}, {V}) = \text{softmax} ({Q} {K}^T) {V}
		\label{eq9}
	\end{equation}
	where the feature-wise similarity between query ${Q}$ and key ${K}$ is measured, and further used to retrieve information value ${V}$ with the computed weight of $\text{softmax}$.
	
	Note that the third dimension of feature volume ${G}_i$ and $\hat{{G}}_i$ is inherently sampled from the epipolar line while traversing all $N_D$ depth values from near to far.
	Along the search space of the epipolar line, we can use self-attention module to retrieve the information in the same view: 
	\begin{equation}
		{H}_i (u,v) = \text{Att} ({G}_i(u,v), {G}_i(u,v), {G}_i(u,v))
		\label{eq10}
	\end{equation}
	where the ${Q}$, ${K}$, ${V}$ of $\text{Att}$ equals to the same feature matrix of epipolar line ${G}_i(u,v) \in \mathbb{R}^{D \times C_1}$.
	In analogy, the trait feature can be calculated: 
	\begin{equation}
		\hat{{H}}_i (u,v) = \text{Att} (\hat{{G}}_i(u,v), \hat{{G}}_i(u,v), \hat{{G}}_i(u,v))
		\label{eq11}
	\end{equation}
	where feature matrix of epipolar line is $\hat{{G}}_i(u,v) \in \mathbb{R}^{D \times C_1}$.

	\noindent\textbf{Epipolar Cross-attention}:
	Note that $i=1$ is the reference view and $i >1$ means the source view.
	The cross-attention on epipolar line among different views can then be calculated:
	\begin{equation}
		{E}_i (u,v) = \text{Att} ({H}_i (u,v), {H}_1 (u,v), {H}_1 (u,v))
		\label{eq12}
	\end{equation}
	where ${Q}$ is the source view feature on view $i$, and ${K}$, ${V}$ equal to the reference view feature ${H}_1 (u,v)$.
	Since the finger trait feature is inherently preserving identity information with reliable correspondence, we further utilize the trait feature as guidance to the cross-attention mechanism:
	\begin{equation}
		\hat{{E}}_i (u,v) = \text{Att} (\hat{{H}}_i (u,v), \hat{{H}}_1 (u,v), {H}_1 (u,v))
		\label{eq13}
	\end{equation}
	where ${Q}$ is source view trait feature $\hat{{H}}_i (u,v)$, ${K}$ is reference view trait feature  $\hat{{H}}_1 (u,v)$, and ${V}$ is the reference view image feature ${H}_1 (u,v)$.
	The cross-view correspondence of trait features is used to retrieve the image feature ${H}_1 (u,v)$.

	\noindent\textbf{Cost Volume Construction}:
	Given the homography warping matrix $\mathcal{H}_i(z)$ estimated by Eq. \ref{eq3}, we can then construct the cost volume by computing the variance of features in the reference view and source view:
	\begin{equation}
		{P}(u,v,z) = \text{Var} ({O}_i (H_i (z) [u,v,1]^T))
		\label{eq14}
	\end{equation}
	where ${O}_i$ is the concatenation of ${E}_i (u,v)$ and $\hat{{E}}_i (u,v)$.
	${P} \in \mathbb{R}^{N_H \times N_W \times N_D \times N_{C1}}$ is the cost volume guided with finger traits.
	
	\noindent\textbf{Radiance Field Construction}:
	The cost volume ${P}$ is fed to a 3D U-Net \cite{chen2021mvsnerf} \textbf{$B$} to aggregate the 3D geometric information.
	\begin{equation}
		{S} = B ({P})
		\label{eq15}
	\end{equation}
	where $S \in \mathbb{R}^{N_H \times N_W \times N_D \times N_{C2}}$ is the output neural encoding volume.
	
	Given an arbitrary 3D location $x$ and a viewing direction $d$, we can then compute the volume density and color as follows:
	\begin{equation}
		\sigma, c_{\text{out}} = A (x, d, S(x), c_{\text{img}}, c_{\text{tra}})
		\label{eq16}
	\end{equation}
	where $c_{\text{img}}$ concatenates the pixel colors of image $I(u_i, v_i)$ from all views, and $c_{\text{tra}}$ concatenates the pixel intensities of binary trait image $\hat{I} (u_i, v_i)$ from all views.
	Here $(u_i, v_i)$ is the pixel coordinate when 3D point $x$ is projected to view $i$.
	$S(x)$ is the trilinearly interpolated feature of neural encoding volume $S$ at the location of 3D point $x$.
	The predicted density $\sigma$ and color $c_{\text{out}}$ construct the volume radiance field on point $x$ and view direction $d$.

	\subsubsection{Training with Extra Geometric Priors}
	\label{methods:fingernerf:extra_priors}
	
	In this section, we first introduce the details of training a generalizable NeRF, and then elaborate the proposed losses which aims to embed the neural rendering of NeRF with extra geometric priors.
	
	\noindent\textbf{End-to-end Training}:
	Given the radiance field $(\sigma, c_{\text{out}})$, the differentiable volume rendering enables the regression of image colors in a discrete form:
	\begin{equation}
		C = \sum_{k} \text{exp} (- \sum_{j=1}^{k-1} \sigma (t_j) \delta_k) (1 - \text{exp} (- \sigma (t_k) \delta_k)) c_{\text{out}} (t_k)
		\label{eq17}
	\end{equation}
	where $\delta_k = t_{k+1} - t_{k}$. 
	Each point along the ray is sampled by a series of depth values $t_k$.
	$C$ is the final rendered pixel color, and $\text{exp} (- \sum_{j=1}^{k-1} \sigma (t_j) \delta_k)$ represents the volume transmittance.
	$k$ is the index of the sampled 3D points along the ray.
	
	The photometric loss for reconstructing realistic pixel colors can be computed as an L2 loss:
	\begin{equation}
		L_{\text{pho}} = \| C - C_{\text{gt}} \|_2^2
		\label{eq18}
	\end{equation}
	where $C_{\text{gt}}$ is the ground truth pixel color in target image.
	This photometric loss is the only loss used to supervised the generalizable NeRF in previous works \cite{chen2021mvsnerf}.
	
	From Eq. \ref{eq17} and \ref{eq18}, we can find that the photometric loss only directly constrains the predicted color $c_{\text{out}}$.
	However, there is no direct entanglement with the volume density $\sigma$, but only the indirect coarse regularization on $\sigma$ based on the color constancy assumption \cite{xu2021learning}.
	The assumption is that the point matches if the color is the same.
	Obviously, such a coarse assumption may fail in finger biometrics, because the images can barely hold abundant variation of color and texture like natural images.
	As a result, this coarse assumption can not provide correct supervision on matching points of different views, and thus misleading the geometry representation in NeRF (shape-radiance ambiguity problem).
	Consequently, it is natural to consider involving extra supervision on the density term $\sigma$ in Eq. \ref{eq17} to handle these problems.
	
	\noindent\textbf{Depth Rendering}:
	To interact with the density $\sigma$, we can render the depth values by simply modifying Eq. \ref{eq17}:
	\begin{equation}
		D = \sum_{k} \text{exp} (- \sum_{j=1}^{k-1} \sigma (t_j) \delta_k) (1 - \text{exp} (- \sigma (t_k) \delta_k)) t_k
		\label{eq19}
	\end{equation}
	where $D$ is the rendered depth value.
	With the help of Eq. \ref{eq19}, we can involve constraints on the rendered depth map to regularize the density $\sigma$.

	\noindent\textbf{Depth Distillation Loss}:
	A naive way is to involve the supervision of ground truth depth maps to Eq. \ref{eq15} like \cite{deng2022depth}.
	However, no available ground truth depth is supported in our task of 3D finger biometrics.
	As an alternative, we distill the pseudo depth ground truth estimated by large model for monocular depth estimation, Midas \cite{ranftl2020towards}, to the depth representation of NeRF.
	Despite of the great performance of Midas on zero-shot depth estimation, the scale of its predicted depth map is totally different from the the scale of 3D fingers in our task.
	Direct supervision with these pseudo depth labels is not feasible because the difference between scales may disturb the training and mislead the training.
	Although the absolute depth is not available for depth supervision, the relative change of depth in local regions can still provide an effective regularization on the 3D shape.

	\begin{figure}[t]%
		\centering
		\includegraphics[width=0.45\textwidth]{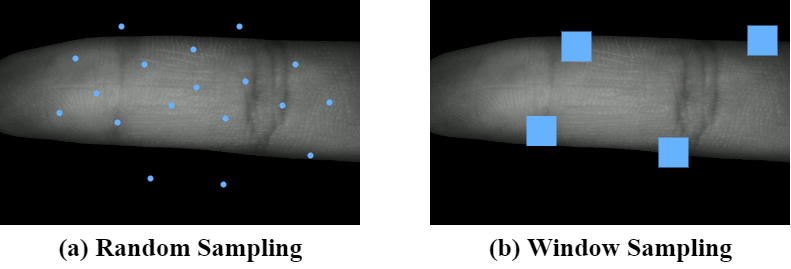}
		\caption{Illustration of the ray sampling strategies in NeRF training.}
		\label{fig:ray_sampling}
		\vspace{-0.4cm}
	\end{figure}
	
	To model the relative change of depth in local regions, we adopt window-based sampling in the neural rendering process.
	As shown in Fig. \ref{fig:ray_sampling}, we randomly sample local windows in images during ray sampling.
	It is used as an alternative to the naive random sampling strategy used in previous NeRFs \cite{chen2021mvsnerf}.
	The rendered color and depth can be concatenated in these windows to local patches.
	
	Assume that there exists scale parameters $ (\theta_s, \theta_t)$ that can map the predicted depth to the same scale of pseudo depth ground truth $D_{\text{pse}}$.
	To align the prediction to the pseudo ground truth, we can optimize these parameters based on a least-squares criterion:
	\begin{equation}
		\theta_s^*, \theta_t^* = \arg \min_{\theta_s, \theta_t} \sum_p (\theta_s D (p) + \theta_t - D_{\text{pse}}(p))^2 
		\label{eq20}
	\end{equation}
	where $\theta_s$ represents the scale and $\theta_t$ represents the shift.
	$p$ is the index of pixel in the sampled window.
	
	Rewriting the formulas into matrix form:
	\begin{equation}
		\theta^* = \arg \min_{\theta} \sum_p ([D (p), 1] \theta - D_{\text{pse}}(p))^2
		\label{eq21}
	\end{equation}
	where $\theta = [\theta_s, \theta_t]^T$.
	
	The closed-form solution is:
	\begin{equation}
		\theta^* = (\sum_p [D (p), 1]^T [D (p), 1])^{-1} (\sum_p [D (p), 1]^T D_{\text{pse}} (p))
		\label{eq22}
	\end{equation}
	The scale parameter $\theta_s^*$ and shift parameter $\theta_t^*$ can be obtained from $\theta^*$.
	
	However, the solution in Eq. \ref{eq22} is not robust to unexpected presence of outliers in pseudo depth labels.
	Specifically, the black background regions in finger images is meaningless and textureless, whose depth map can not be correctly estimated by any method and leads to an incorrect solution in Eq. \ref{eq22}.
	Hence, we further modify Eq. \ref{eq22} as follows:
	\begin{equation}
		\theta^* = (\sum_p [D (\hat{p}), 1]^T [D (\hat{p}), 1])^{-1} (\sum_p [D (\hat{p}), 1]^T D_{\text{pse}} (p))
		\label{eq23}
	\end{equation}
	where $\hat{p}$ is the index of meaningful pixel in the sampled windows.
	All pixels located in the black background regions are abandoned.
	
	Furthermore, we treat $\theta_s$ and $\theta_t$ as learnable parameters in the training phase.
	The solution $\theta^*$ of Eq. \ref{eq23} is used as of $\theta$ an initialization during training.
	The mean-squared loss can be formulated to supervised the predicted depth maps with pseudo depth labels:
	\begin{equation}
		L_{\text{dep}} = \| \theta_s D + \theta_t - D_{\text{pse}} \|_2^2
		\label{eq24}
	\end{equation}

	\noindent\textbf{Trait Guided Rendering Loss}:
	As the finger traits naturally preserves identity-preserving visual clues, the correspondence between the trait image on different views can be used to involve extra matching priors in the neural rendering.
	The original photometric neural rendering loss in Eq. \ref{eq14} is easy to overfit and converge to the mean values of pixels in local regions, thus suffering from oversmoothing effect.
	Following the same ray sampling strategy of window based sampling shown in Fig. \ref{fig:ray_sampling}, we utilize the trait intensity in the sampled window as a soft weight to determine which regions to foucs on
	It can enforce the neural rendering loss to concentrate on the areas with distinguishing trait features, benefiting the optimization of multi-view correspondence.
	
	Concretely, the binary trait intensities $C_{\text{tra}} \in \mathbb{R}^{s \times s}$ in the sampled windows is firstly smoothed by the softmax function:
	\begin{equation}
		w_{\text{tra}} = \text{softmax}_{p} (C_{\text{tra}})
		\label{eq25}
	\end{equation}
	where $p$ is the index of pixels in the $s \times s$ window.
	The original binary trait intensity is converted to a soft score by $\text{softmax}$ function.
	
	Then the frist-order rendering loss can be computed via:
	\begin{equation}
		L_{\text{1st}} = \| w_{\text{tra}} \odot (C - C_{\text{gt}}) \|_2^2
		\label{eq26}
	\end{equation}
	
	The second-order rendering loss can be computed by:
	\begin{equation}
		\begin{aligned}
			L_{\text{2nd}} = &\| w_{\text{tra}} \odot (\nabla_x C -  \nabla_x C_{\text{gt}}) \|_2^2 + \\ &\| w_{\text{tra}} \odot (\nabla_y C -  \nabla_y C_{\text{gt}}) \|_2^2
			\label{eq27}
		\end{aligned}
	\end{equation}
	where $\nabla_x$ means the gradient along $x$ axis of the image, and $\nabla_y$ means the gradient along $y$ axis of the image.
	
	The trait guided rendering loss can then be formulated as:
	\begin{equation}
		L_{\text{tra}} = L_{\text{1st}} + L_{\text{2nd}}
		\label{eq28}
	\end{equation}

	\noindent\textbf{Overall Loss}:
	Finally, the overall loss can be computed:
	\begin{equation}
		L = \lambda_1 L_{\text{tra}} + \lambda_2 L_{\text{dep}}
		\label{eq29}
	\end{equation}
	where $\lambda_1 = 1.0$, $\lambda_2=0.1$ in default.
	
	\begin{table*}[t]
		\centering
		\caption{Brief introduction of utilized contactless 3D finger trait datasets in this paper.}
		\resizebox{\linewidth}{!}{
			\begin{tabular}[H]{cccc}
				\textbf{Dataset} & \textbf{SCUT-Finger-3D} & \textbf{SCUT-Fingervein-3D} & \textbf{UNSW-3D} \\ \hline
				Sample      & \begin{minipage}[b]{0.6\columnwidth}
					\centering
					\vspace{0.1cm}
					\raisebox{-.5\height}{\includegraphics[width=\linewidth]{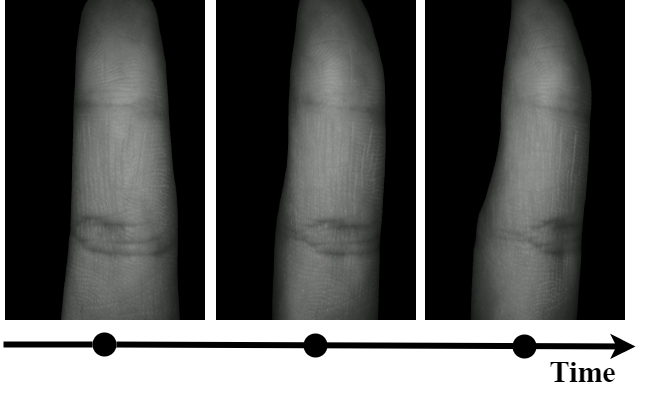}}
				\end{minipage}
				& \begin{minipage}[b]{0.6\columnwidth}
					\centering
					\vspace{0.1cm}
					\raisebox{-.5\height}{\includegraphics[width=\linewidth]{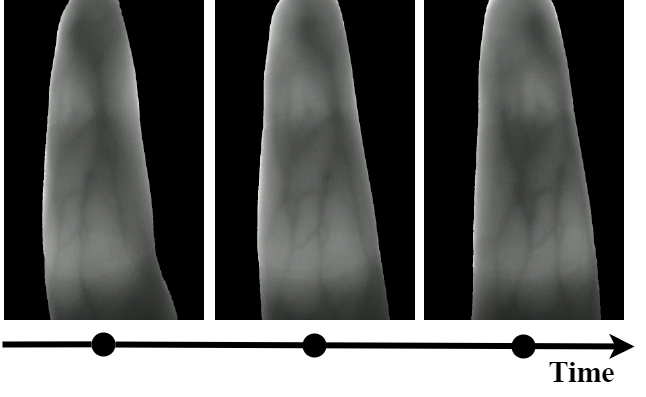}}
				\end{minipage}               
				& \begin{minipage}[b]{0.6\columnwidth}
					\centering
					\vspace{0.1cm}
					\raisebox{-.5\height}{\includegraphics[width=\linewidth]{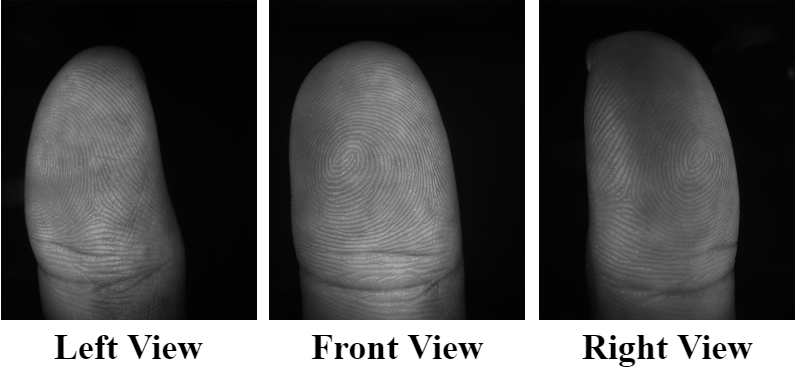}}
				\end{minipage}        \\ \hline
				Data type   & \makecell{Contactless finger images}    &  \makecell{Contactless fingervein images}                  & \makecell{Contactless finger images}        \\ \hline
				Cameras     & 1 & 1 & 3 \\ \hline
				Description & \makecell{Video sequences of 660 fingers; \\ Volunteers are asked to rotate their \\ finger from -30 degrees to 30 degrees.}               &  \makecell{Video sequences of 660 fingers; \\ Volunteers are asked to rotate their \\ finger from -30 degrees to 30 degrees.}                  &  \makecell{3 view images of 1500 fingers.}       \\ \hline
				Data Split  & train : valid : test = 5 : 2 : 3 & train : valid : test = 5 : 2 : 3 & train : valid : test = 5 : 2 : 3 \\ \hline
		\end{tabular}}
		
		\label{tab:database_introcution}
		\vspace{-0.3cm}
	\end{table*}
	
	\section{Experiment}
	\label{experiment}

	In this section, we introduce the implementation details and benchmark evaluations respectively in Section \ref{experiment:implementation} and \ref{experiment:benchmark}.
	Then we conduct detailed experiments aimed at addressing the following questions:
	\begin{itemize}
		\item \textbf{Q1}: Aiming at boosting the finger trait recognition with 3D information, \emph{how does our implicit method compared with the existing explicit 3D reconstruction pipelines}? (Section \ref{experiment:implicit_vs_explicit})
		\item \textbf{Q2}: Considering the one-shot setting in open-set finger trait recognition, the neural rendering techniques should be generalizable without per-scene finetuning. \emph{How does our proposed method compared with other NeRF-based techniques under the generalizable neural rendering setting?} (Section \ref{experiment:generalizable_rendering})
		\item \textbf{Q3}: As a trade-off between efficiency and accuracy in 3D finger trait recognition, the multi-view methods with sparse views lead the state-of-the-art. \emph{Can the proposed method bring some improvement compared with these methods?} (Section \ref{experiment:multi_view})
		\item \textbf{Q4}:Existing 3D finger recognition methods are highly customized towards specific device capturing single modality like fingerprint image or finger vein image. \emph{Can the proposed methods be generalized to different modalities of finger traits like fingerprint or finger vein images?} (Experiments under different modalities in Section \ref{experiment:implicit_vs_explicit}, \ref{experiment:generalizable_rendering}, and \ref{experiment:multi_view}.)
		
	\end{itemize}
	
	\subsection{Dataset}
	
	As summarized in Table \ref{tab:database_introcution}, three databases are used for evaluating the proposed method: SCUT-Finger-3D, SCUT-FingerVein-3D, UNSW-3D.
	Due to the particularity of the neural rendering task in this paper, we propose two novel datasets for neural rendering of finger biometrics: SCUT-Finger-3D and SCUT-FingerVein-3D.
	These two datasets are used to evaluate the generalization ability of the proposed method towards different modalities of finger traits.
	UNSW-3D is a publicly available database which is widely used in 3D fingerprint.
	We utilize UNSW-3D to evaluate the generalization performance of the proposed method.

	\textbf{SCUT-Finger-3D} contains contactless finger images.
	The camera is set in the bottom of the imaging device.
	Volunteers are asked to rotate their finger from -30 degrees to 30 degrees causually, meantime the video is captured to represent a finger.
	Assume that the finger is rigid, different frames of the video can be treated as multi-view images.
	We use this dataset to evaluate the generalization ability of the proposed FingerNeRF.
	
	\textbf{SCUT-FingerVein-3D} contains contactless finger vein images captured under infrared light.
	In analogy with SCUT-Finger-3D, the camera is set in the bottom of the imaging device.
	The only difference compared with SCUT-Finger-3D is that this dataset is based on a different modality of finger vein images.
	We use this dataset to evaluate the generalization ability of the proposed FingerNeRF across different modalities.
	
	\textbf{UNSW-3D} \cite{lin2018matching} also contains contactless finger images with clear fingerprints. We use the raw finger images of this dataset to evaluate the generalization performance of the proposed method towards other multi-view datasets.

	\subsection{Implementation Details}
	\label{experiment:implementation}
	
	The experiments are conducted with one RTX 3090 GPU.
	The output feature dimension of TGT is $N_{C1}=8$.
	We construct the cost volume of multi-view images with $N_{C2}=8$ channels, which is also the output dimension of the feature extraction network with shared weights.
	During the ray sampling process, we adopt $N_d=128$ depth hypotheses uniformly sampled from the nearest to the farthest depth planes.
	The sampled window size is set to $s \times s = 64 \times 64$ in default.
	We train the FingerNeRF on each dataset for 20 epochs, which may take half a day in total.
	The resolution of image is resized to $N_H \times N_W = 320 \times 200$ during training in order to remedy the cost of GPU memory.
	
	\begin{table*}[t]
		\centering
		\caption{Implicit vs explicit 3D recognition pipelines with only 3D geometric modality on SCUT-Finger-3D dataset. Note that our implicit method only requires 3 views randomly selected from the video.}
		\resizebox{0.9\linewidth}{!}{
			\begin{tabular}{l|c|c|cccccc}
				\hline
				\rowcolor{gray!40}
				\textbf{Method}                      & \textbf{Input Views}   & \textbf{Modality} & \textbf{EER} $^\downarrow$ & \textbf{T-F=0.01} $^\uparrow$     & \textbf{T-F=0.001} $^\uparrow$     & \textbf{mAP} $^\uparrow$     & \textbf{Rank 1} $^\uparrow$  & \textbf{Rank 5} $^\uparrow$  \\ \hline
				PointNet \cite{qi2017pointnet}               & Dense  & Points & 37.85\% & 19.15\% & 16.15\% & 25.84\% & 26.44\% & 40.47\%      \\
				PointNet++ \cite{qi2017pointnet++}           & Dense  & Points & 37.45\% & 23.30\% & 19.52\% & 28.51\% & 27.61\% & 44.83\%      \\
				DGCNN \cite{wang2019dynamic}                 & Dense  & Points & 36.81\% & 22.09\% & 16.67\% & 29.15\% & 30.08\% & 43.86\%       \\
				DPAM \cite{liu2019dynamic}                   & Dense  & Points & 37.11\% & 21.69\% & 16.11\% & 38.39\% & 28.97\% & 43.38\%        \\
				GSNet \cite{xu2020geometry}                 & Dense  & Points & 40.93\% & 8.99\%  & 4.20\% & 14.92\% & 18.05\%	 & 38.28\%         \\
				GBNet \cite{qiu2021geometric}              & Dense  & Points & 39.03\% & 11.00\% & 5.60\% & 21.52\% & 19.65\% & 37.46\%       \\
				GDANet \cite{xu2021learning}                 & Dense  & Points & 38.26\% & 23.24\% & 19.85\% & 28.31\% & 26.25\% & 40.32\%         \\
				PointMLP-E \cite{ma2022rethinking}           & Dense  & Points & 36.74\% & 21.21\% & 16.09\% & 22.53\% & 25.08\% & 36.69\%         \\
				PointMLP \cite{ma2022rethinking}             & Dense  & Points & 38.80\% & 20.21\% & 16.73\% & 20.32\% & 22.99\% & 36.92\%         \\ 
				Point-Transformers-H \cite{zhao2021point}    & Dense  & Points & 37.02\% & 19.30\% & 12.85\% & 23.78\% & 24.94\% & 41.97\%        \\
				Point-Transformers-M \cite{engel2021point}   & Dense  & Points & 37.82\% & 22.64\% & 18.76\% & 23.81\% & 24.21\% & 37.80\%         \\ 
				Point-Transformers-N \cite{guo2021pct}       & Dense  & Points & 40.64\% & 17.88\% & 9.58\% & 16.84\% & 13.68\% & 38.57\%        \\ 
				\hline
				FingerNeRF (Ours)                                   & Sparse(3)   & Depths & \textbf{22.60\%} & \textbf{28.60\%} & \textbf{25.00\%} & \textbf{45.98\%} & \textbf{36.30\%} & \textbf{58.67\%}     \\
				\hline
			\end{tabular}
		}
		\label{tab:implicit_vs_explicit_geo_only}
		\vspace{-0.6cm}
	\end{table*}
	
	\begin{table*}[t]
		\centering
		\caption{Implicit vs explicit 3D recognition pipelines with both 3D geometric and texture modality on SCUT-Finger-3D dataset. Note that our implicit method only requires 3 views randomly selected from the video.}
		\resizebox{0.9\linewidth}{!}{
			\begin{tabular}{l|c|c|cccccc}
					\hline
					\rowcolor{gray!40}
					\textbf{Method}                      & \textbf{Input Views}   & \textbf{Modality} & \textbf{EER} $^\downarrow$ & \textbf{T-F=0.01} $^\uparrow$     & \textbf{T-F=0.001} $^\uparrow$     & \textbf{mAP} $^\uparrow$     & \textbf{Rank 1} $^\uparrow$  & \textbf{Rank 5} $^\uparrow$  \\ \hline
					PointNet \cite{qi2017pointnet}                   & Dense         & Points + Color   & 23.81\% & 26.76\% & 21.55\% & 34.20\% & 33.72\% & 61.67\%      \\
					PointNet++ \cite{qi2017pointnet++}                 & Dense         & Points + Color   & 20.23\% & 29.97\% & 19.76\% & 34.60\% & 38.09\% & 64.92\%      \\
					DGCNN \cite{wang2019dynamic}                      & Dense         & Points + Color   & 16.14\% & 33.70\% & 23.73\% & 45.51\% & 37.60\% & 64.24\%        \\
					DPAM  \cite{liu2019dynamic}                      & Dense         & Points + Color   & 17.91\% & 28.00\% & 0.00\% & 35.95\% & 39.93\% & 63.61\%       \\
					GSNet \cite{xu2020geometry}                     & Dense         & Points + Color   & 16.41\% & 25.79\% & 12.88\% & 34.59\% & 39.01\% & 67.63\%        \\
					GBNet \cite{qiu2021geometric}                    & Dense         & Points + Color   & 18.64\% & 32.76\% & 23.03\% & 38.12\% & 37.80\% & 66.52\%       \\
					GDANet \cite{xu2021learning}                     & Dense         & Points + Color   &  18.99\% & 32.30\% & 24.06\% & 39.59\% & 38.91\% & 69.87\%        \\
					PointMLP-E \cite{ma2022rethinking}                 & Dense         & Points + Color   & 27.27\% & 16.73\% & 4.45\% & 20.97\% & 20.28\% & 50.70\%        \\
					PointMLP \cite{ma2022rethinking}                   & Dense         & Points + Color   & 26.39\% & 12.42\% & 2.85\% & 20.77\% & 20.67\% & 51.48\%        \\ 
					Point-Transformers-H \cite{zhao2021point}       & Dense         & Points + Color   & 17.71\% & 31.58\% & 10.96\% & 32.21\% & 35.52\% & \textbf{71.86\%}       \\
					Point-Transformers-M \cite{engel2021point}       & Dense         & Points + Color   & 17.45\% & 32.24\% & 17.21\% & 35.73\% & 38.18\% & 69.57\%        \\ 
					Point-Transformers-N \cite{guo2021pct}       & Dense         & Points + Color   & 40.90\% & 14.21\% & 5.42\% & 14.84\% & 15.57\% & 39.50\%       \\ 
					\hline
					FingerNeRF (Ours)                        & Sparse(3)   & Depths + Color   & \textbf{15.60\%} & \textbf{42.00\%} & \textbf{29.00\%} & \textbf{53.97\%} & \textbf{41.93\%} & 68.00\%       \\
					\hline
				\end{tabular}
			}
			\label{tab:implicit_vs_explicit_geo_tex}
			\vspace{-0.4cm}
		\end{table*}
		
		\begin{figure*}
			\centering
			\begin{subfigure}{0.45\textwidth}
				\centering
				\includegraphics[width=\textwidth]{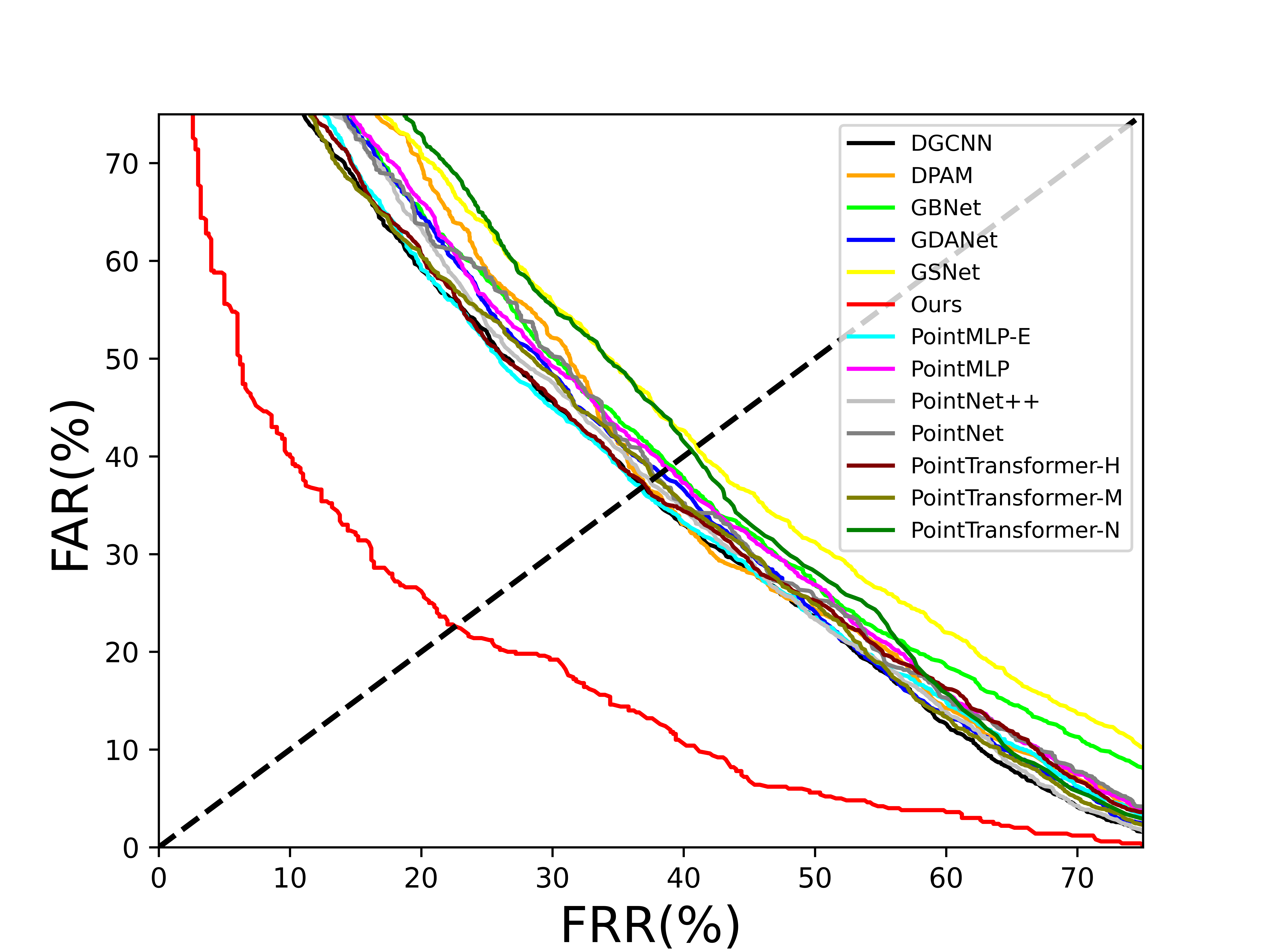}
				\caption{Geometric}
				\label{fig:implicit_vs_explicit_roc_geo_only}
			\end{subfigure}
			\hfill
			\begin{subfigure}{0.45\textwidth}
				\centering
				\includegraphics[width=\textwidth]{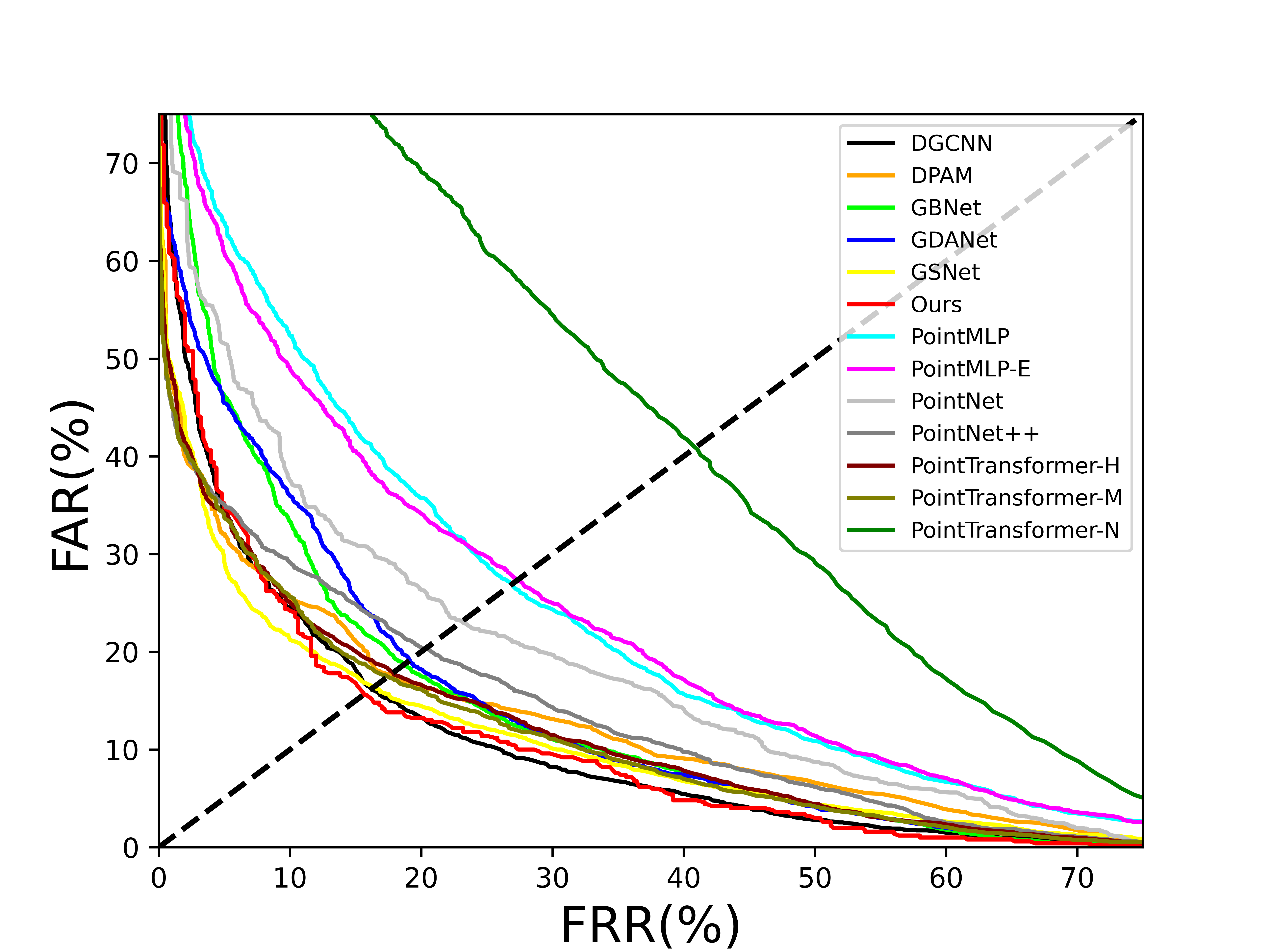}
				\caption{Geometric + Texture}
				\label{fig:implicit_vs_explicit_roc_geo_tex}
			\end{subfigure}
			\caption{The Detection Error Tradeoff (DET) curves on SCUT-Finger-3D dataset.}
			\label{fig:implicit_vs_explicit_roc}
			\vspace{-0.4cm}
		\end{figure*}
		
		\subsection{Benchmark Evaluation}
		\label{experiment:benchmark}
		
		\subsubsection{Protocol and Task}
		
		For data-driven biometric recognition methods, the \textbf{subject-dependent protocol} assumes that all identities in the test set are the predefined in the training set.
		This ideal setting rarely fits the real-world scenario, because it is more likely to happen that the evaluation process and the training process may have samples with different identities, which follows the \textbf{subject-independent protocol} \cite{wang2021deep}.
		Since the testing identities are disjoint from the training identities, the evaluation metrics can demonstrate the open-set generalization performance of the methods in a better way.

		Biometric benchmarks usually have two kinds of tasks: verification and identification.
		Verification task computes the 1-to-1 similarity score between a probe and a gallery to find out whether the two samples are from the same subject.
		Identification task \cite{jain201650} has two different settings: close-set identification and open-set identification.
		The close-set setting requires the probe to appear in the gallery identities, while in the open-set setting, the probe may not appear in the gallery identities.
		Following \cite{yang2021lfmb}, we conduct experiments on the the \textbf{verification} task and \textbf{close-set identification} task following the \textbf{subject-independent protocol} and \textbf{subject-dependent protocol} respectively.
		
		\subsubsection{Dataset Division}
		
		In our experiments, each finger is considered as an individual subject.
		We randomly separate all subjects of the whole dataset into \textbf{training set}, \textbf{validation set} and \textbf{testing set} with the ratio of \textbf{5:2:3}.
		Following \cite{yang2021lfmb}, we \textbf{fix the partition} of the separated sets and record the subject names into csv files of the corresponding identities to ensure the reproducibility and fair comparison in this work.
		
		\subsubsection{Evaluation Metrics}
		
		For the verification task, the adopted metrics are \textbf{Detection Error Tradeoff (DET) curve}, \textbf{Equal Error Rate (EER)}, \textbf{TAR@FAR=0.01}, and \textbf{TAR@FAR=0.001}.
		DET curve is ploted to reflect the variation of Fasle Accept Rate (FAR) versus False Reject Rate (FRR) when varying the threshold of the matching scores between different samples.
		It also represents the tradeoff between the metric of FAR and FRR.
		EER is the value when FAR and FRR are equal, which is an overall measure of performance in biometrics.
		TAR@FAR=0.01 and TAR@FAR=0.001 are measures of practical scenarios \cite{yang2021lfmb}.
		TAR@FAR=0.01 is the True Accept Rate (TAR) when the False Accept Rate (FAR) equals 0.01.
		In analogy, TAR@FAR=0.001 reflects the TAR when FAR is 0.001.
		
		For the identification task, the adopted metrics are \textbf{Mean Average Precision (mAP)}, \textbf{Rank-1 Accuracy} and \textbf{Rank-5 Accuracy} in the \textbf{Cumulative Match Characteristic (CMC) curve}.
		CMC curve plots the probability that the positive result can be found in the top K samples sorted with the matching scores.
		The horizontal axis is K, and the vertical axis is the prbability of positive samples.
		Rank-1 and Rank-5 Accuracy are the vertical value of CMC when K is 1 and 5 respectively.
		These metrics are important metrics in practical identification tasks.
		Besides, to evaluate the overall representative of performance, the metric of mAP is also used here to measure the overall ranking effect.
		
		\subsection{Implicit vs Explicit 3D Reconstruction for Finger Trait Recogntion}
		\label{experiment:implicit_vs_explicit}
		
		\subsubsection{Experiment Settings}
		\label{experiment:implicit_vs_explicit:setting}
		
		In this section, we aim to evaluate the effectiveness of the reconstructed 3D representation in finger trait recognition tasks.
		These reconstruction methods can be categorized in to two categories: explicit methods, and our proposed implicit methods.
		
		\begin{table*}[t]
			\centering
			\vspace{-0.4cm}
			\caption{Comparison of explicit 3D recognition methods and our implicit method with geometric modality on SCUT-FingerVein-3D dataset.}
			\label{tab:implicit_vs_explicit_geo_only_fv}
			\resizebox{0.9\linewidth}{!}{
				\begin{tabular}{l|c|c|cccccc}
					\hline
					\rowcolor{gray!40}
					\textbf{Method}                      & \textbf{Input Views}   & \textbf{Modality} & \textbf{EER} $^\downarrow$ & \textbf{T-F=0.01} $^\uparrow$     & \textbf{T-F=0.001} $^\uparrow$     & \textbf{mAP} $^\uparrow$     & \textbf{Rank 1} $^\uparrow$  & \textbf{Rank 5} $^\uparrow$  \\ \hline
					PointNet \cite{qi2017pointnet}               & Dense  & Points & 38.76\% & 17.62\% & 13.50\% & 24.70\% & 25.42\% & 44.57\%       \\
					PointNet++ \cite{qi2017pointnet++}           & Dense  & Points & 38.40\% & 19.82\% & 0.00\% & 22.31\% & 22.59\% & 40.42\%      \\
					DGCNN \cite{wang2019dynamic}                 & Dense  & Points & 37.39\% & 19.41\% & 14.44\% & 27.24\% & 24.76\% & 44.95\%      \\
					DPAM \cite{liu2019dynamic}                   & Dense  & Points & 37.96\% & 19.71\% & 14.82\% & 22.06\% & 23.73\% & 40.28\%       \\
					GSNet \cite{xu2020geometry}                  & Dense  & Points & 37.85\% & 19.18\% & 14.59\% & 22.09\% & 23.16\% & 42.92\%       \\
					GBNet \cite{qiu2021geometric}                & Dense  & Points & 37.38\% & 14.29\% & 0.00\% & 21.22\% & 22.21\% & 37.02\%     \\
					GDANet \cite{xu2021learning}                 & Dense  & Points & 36.30\% & 18.18\% & 14.03\% & 24.74\% & 23.68\% & 44.06\%      \\
					PointMLP-E \cite{ma2022rethinking}           & Dense  & Points & 38.51\% & 17.47\% & 13.73\% & 22.35\% & 22.17\% & 40.42\%      \\
					PointMLP \cite{ma2022rethinking}             & Dense  & Points & 37.74\% & 19.94\% & 13.35\% & 25.05\% & 24.43\% & 40.24\%     \\ 
					Point-Transformers-H \cite{zhao2021point}    & Dense  & Points & 36.95\% & 18.73\% & 12.79\% & 20.45\% & 25.09\% & 37.26\%     \\
					Point-Transformers-M \cite{engel2021point}   & Dense  & Points & 36.85\% & 19.76\% & 14.64\% & 24.54\% & 24.62\% & 39.19\%      \\
					Point-Transformers-N \cite{guo2021pct}       & Dense  & Points & 41.01\% & 20.32\% & 14.88\% & 24.37\% & 23.63\% & 40.14\%     \\ 
					\hline
					FingerNeRF (Ours)                                    & Sparse(3)   & Depths & \textbf{25.28\%} & \textbf{25.09\%} & \textbf{19.62\%} & \textbf{36.91\%} & \textbf{27.51} & \textbf{56.61\%}      \\
					\hline
				\end{tabular}
			}
			\vspace{-0.4cm}
		\end{table*}

		\begin{table*}[t]
			\centering
			\caption{Comparison of explicit 3D recognition methods and our implicit method with geometric and texture modality on SCUT-FingerVein-3D dataset.}
			\resizebox{0.9\linewidth}{!}{
				\begin{tabular}{l|c|c|cccccc}
						\hline
						\rowcolor{gray!40}
						\textbf{Method}                      & \textbf{Input Views}   & \textbf{Modality} & \textbf{EER} $^\downarrow$ & \textbf{T-F=0.01} $^\uparrow$     & \textbf{T-F=0.001} $^\uparrow$     & \textbf{mAP} $^\uparrow$     & \textbf{Rank 1} $^\uparrow$  & \textbf{Rank 5} $^\uparrow$  \\ \hline
						PointNet \cite{qi2017pointnet}               & Dense & Points + Color   &34.69\% & 18.69\% & 14.52\% & 26.48\% & 22.07\% & 46.04\%       \\
						PointNet++ \cite{qi2017pointnet++}           & Dense & Points + Color   &33.46\% & 17.56\% & 10.85\% & 25.07\% & 23.21\% & 49.53\%       \\
						DGCNN \cite{wang2019dynamic}                 & Dense & Points + Color   &32.25\% & 21.21\% & 12.32\% & 29.15\% & 25.42\% & 46.98\%    \\
						DPAM  \cite{liu2019dynamic}                  & Dense & Points + Color   &35.72\% & 16.47\% & 0.00\% & 28.11\% & 23.20\% & 41.56\%      \\
						GSNet \cite{xu2020geometry}                  & Dense & Points + Color   &33.54\% & 19.12\% & 13.14\% & 28.61\% & 27.74\% & 45.66\%       \\
						GBNet \cite{qiu2021geometric}                & Dense & Points + Color   &32.97\% & 18.91\% & 13.18\% & 21.43\% & 23.54\% & 49.86\%     \\
						GDANet \cite{xu2021learning}                 & Dense & Points + Color   &30.35\% & 18.59\% & 13.53\% & 29.01\% & 25.66\% & 49.72\%    \\
						PointMLP-E \cite{ma2022rethinking}           & Dense & Points + Color   &33.60\% & 18.94\% & 13.29\% & 23.16\% & 23.34\% & 44.66\%      \\
						PointMLP \cite{ma2022rethinking}             & Dense & Points + Color   &32.13\% & 19.91\% & 13.65\% & 28.25\% & 25.61\% & 49.20\%      \\
						Point-Transformers-H \cite{zhao2021point}    & Dense & Points + Color   &32.56\% & 18.92\% & 11.44\% & 28.58\% & 25.05\% & 45.28\%      \\
						Point-Transformers-M \cite{engel2021point}   & Dense & Points + Color   &32.06\% & 19.85\% & 13.24\% & 27.25\% & 23.92\% & 45.99\%      \\
						Point-Transformers-N \cite{guo2021pct}       & Dense & Points + Color   &40.54\% & 20.32\% & 16.02\% & 24.22\% & 21.13\% & 35.75\%      \\
						\hline
						FingerNeRF (Ours)                        & Sparse(3)   & Depths + Color   & \textbf{16.98\%} & \textbf{28.87\%} & \textbf{26.98\%} & \textbf{42.38\%} & \textbf{31.35\%} & \textbf{65.34\%}      \\
						\hline
					\end{tabular}
				}
				\label{tab:implicit_vs_explicit_geo_tex_fv}
				\vspace{-0.4cm}
			\end{table*}

			\begin{figure*}
				\centering
				\begin{subfigure}{0.45\textwidth}
					\centering
					\includegraphics[width=\textwidth]{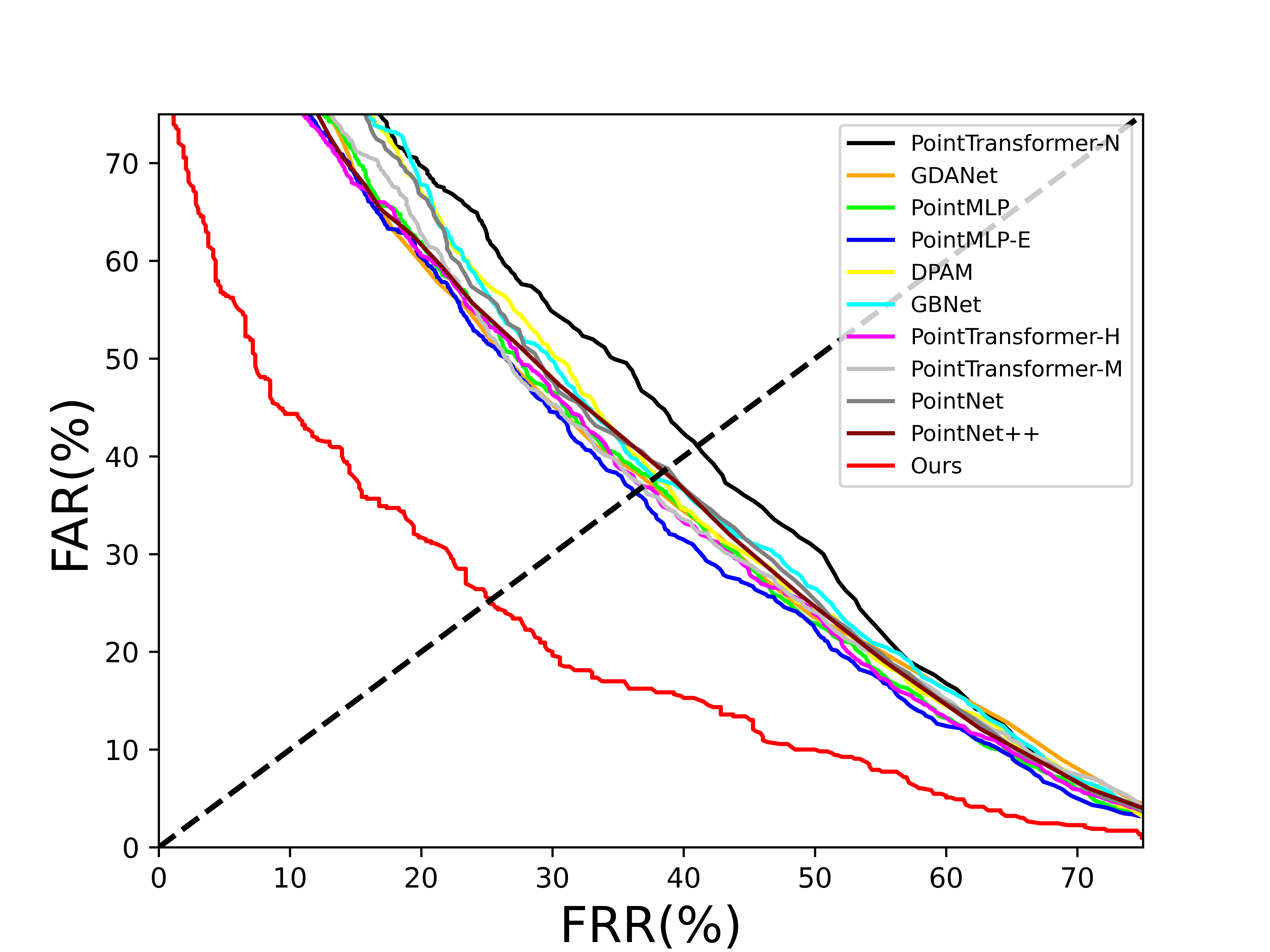}
					\caption{Geometric}
					\label{fig:implicit_vs_explicit_roc_geo_only_fv}
				\end{subfigure}
				\hfill
				\begin{subfigure}{0.45\textwidth}
					\centering
					\includegraphics[width=\textwidth]{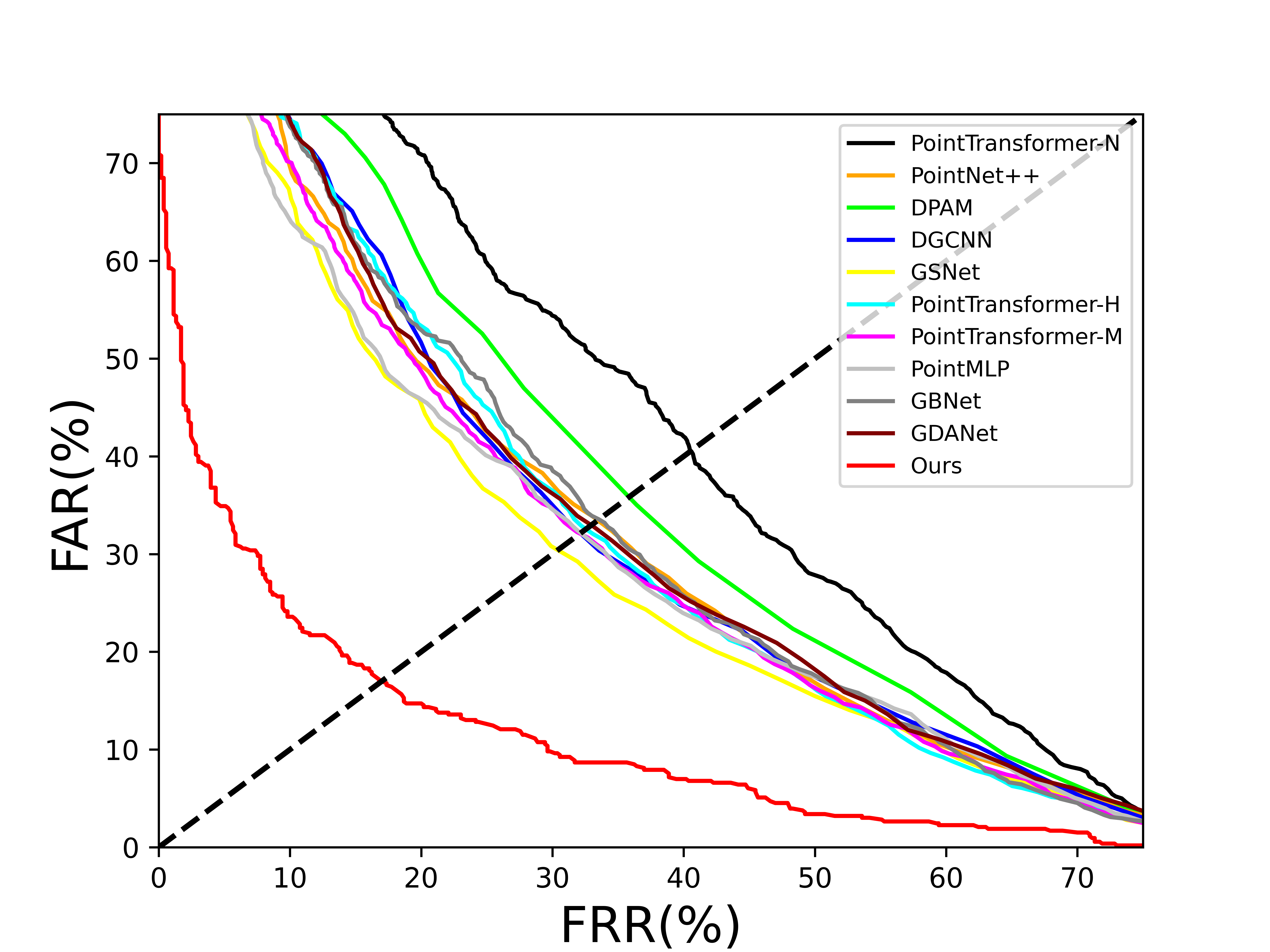}
					\caption{Geometric + Texture}
					\label{fig:implicit_vs_explicit_roc_geo_tex_fv}
				\end{subfigure}
				\caption{The Detection Error Tradeoff (DET) curves on SCUT-FingerVein-3D dataset.}
				\label{fig:implicit_vs_explicit_roc_fv}
				\vspace{-0.4cm}
			\end{figure*}
			
			\begin{enumerate}
				\item For \textbf{explicit methods}, the pipeline is comprised of two steps: firstly reconstruct the 3D model from given images, and then extract 3D features from the reconstructed models. 
				We adopt the benchmarking software for 3D reconstruction, COLMAP \cite{schoenberger2016sfm, schoenberger2016mvs}, to reconstruct the 3D point clouds from the video sequences explicitly. 
				\textbf{The explicit reconstruction process requires dense view inputs containing all available images in the video sequences}.
				The reconstructed point clouds are further fed to various networks specially designed for point cloud perception for feature extraction, including: PointNet \cite{qi2017pointnet}, PointNet++ \cite{qi2017pointnet++}, DGCNN \cite{wang2019dynamic}, DPAM  \cite{liu2019dynamic}, GSNet \cite{xu2020geometry}, PointMLP-E \cite{ma2022rethinking}, PointMLP-E/PointMLP \cite{ma2022rethinking}, Point-Transformers-H \cite{zhao2021point}, Point-Transformers-M \cite{engel2021point}, and Point-Transformers-N \cite{guo2021pct}.
				
				\item For our \textbf{implicit method}, our FingerNeRF requires several multi-view images as input and can render color images and depth maps on arbitrary views given the camera poses.
				Instead of utilizing dense view inputs like explicit methods, \textbf{we adopt a more challenging setting with sparse view inputs (3 views in default)}.
				We randomly select 3 views from all videos in the dataset to represent the 3D subject, and train the FingerNeRF to reconstruct the remaining views in the video given only 3 views as input.
				After training, the FingerNeRF is used to render $V$ unseen views randomly selected from the camera trajectory of the video to construct multi-view RGB and depth maps.
				Note that $V$ is set to 20 in default.
				Then these multi-view RGB and depth maps are further fed to a multi-view convolution network \cite{su2015multi} to extract the feature embedding for finger trait recognition.
				
			\end{enumerate}

			The comparison between implicit and explicit pipelines can reflect the effectiveness of the reconstructed 3D finger traits for recognition.
			Furthermore, we conduct the experiments under different modalities respectively in SCUT-Finger-3D (Section \ref{experiment:implicit_vs_explicit:scut_finger_3d}) and SCUT-FingeVein-3D (Section \ref{experiment:implicit_vs_explicit:scut_fingervein_3d}).
			The ablation experiments are provided in Section \ref{experiment:implicit_vs_explicit:ablation} to evaluate the effectiveness of different components.

			\subsubsection{Comparison on SCUT-Finger-3D}
			\label{experiment:implicit_vs_explicit:scut_finger_3d}
			
			In this section, we provide the qualitative results of finger trait recognition in Table \ref{tab:implicit_vs_explicit_geo_only} and \ref{tab:implicit_vs_explicit_geo_tex}.
			Both of the verification and identification tasks are evaluated and the results of explicit and our implicit 3D methods are compared in the tables.
			Table \ref{tab:implicit_vs_explicit_geo_only} only utilizes the 3D geometric feature to evaluate the effectiveness of 3D shape information reconstructed from given inputs in the aforementioned verification and identification tasks.
			Table \ref{tab:implicit_vs_explicit_geo_tex} reports the experimental results when both 3D geometric feature and texture features are used in verification and identification tasks.
			Moreover, we provide the visualization of DET curves of finger trait recognition with different modalities in Fig. \ref{fig:implicit_vs_explicit_roc}.
			
			In Table \ref{tab:implicit_vs_explicit_geo_only}, we only utilize the 3D geometric feature for finger trait recognition and compare the performance with explicit methods.
			The explicit methods only use the X/Y/Z coordinates of the reconstructed 3D point cloud in the experiments, and our implicit methods only utilizes the rendered multi-view depth maps for comparison.
			As shown in the table, the EER of FingerNeRF is 22.60\% which is much better than the EERs of explicit methods that is about 35\% to 40\%.
			Other verification metrics like T-F=0.01 and T-F=0.001 also reflect impressive improvements compared with explicit methods.
			For the identification metrics, the mAP of our method is 45.98\% which performs better when compared with the ones of explicit methods which is about 15\% to 38\%.
			Other identification metrics like Rank-1 and Rank-5 also show that our method performs better than the ones of explicit methods.
			These results proves that the proposed implicit methods contains much more effective 3D information for finger trait recognition task compared with the explicit method.
			Note that our FingerNeRF only adopts 3 views randomly selected from the video rather the whole sequence required in the explicit 3D reconstruction.
			The reason that our FingerNeRF outperforms explicit methods can be resorted to two reasons:
			\begin{itemize}
				\item On the one hand, the explicit methods depends on representative feature descriptors like SIFT \cite{lowe1999object} and ORB \cite{rublee2011orb}, whereas these feature descriptors are too sparse to cover the whole scene on the specific domain of the captured finger images. Due to the strict geometric checking process based on point reprojection error among views in explicit methods, many useful geometric structures of the scene are lossed.
				\item On the other hand, our implicit method utilizes neural network to extract the feature descriptors which is trained self-supervisedly via differentiable neural rendering on this specific domain. 
				The designed training process can fit the learnable feature descriptors to this specific domain and ensure the dense correspondence among views recovering much more accurate and complete 3D information than explicit methods.
			\end{itemize}

			In Table \ref{tab:implicit_vs_explicit_geo_tex}, both the geometric and texture modalities are utilized in the verification and identification experiments.
			For explicit methods, each point of the reconstructed point clouds is embedded with X/Y/Z coordinates and the R/G/B values on corresponding pixel.
			The input dimension of utilized point cloud perception networks are modified to 6 accordingly.
			For our implicit method, we utilize both the rendered color images and depth maps for experiments.
			As the table shows, the EER of FingerNeRF is 15.60\% that performs slightly better compared with other explicit methods.
			On other verification metrics like  and T-F=0.0001, our proposed method is significantly better than other explicit methods, respectively achieving an improvement of 8.3\% on T-F=0.01 and 4.94\% on T-F=0.001.
			On the identification metrics, our FingerNeRF can achieve an improvement of 8.46\% on mAP and 2.00\% on Rank 1.
			
			In Fig. \ref{fig:implicit_vs_explicit_roc}, the DET curve of aforementioned experiments are visualized.
			Fig. \ref{fig:implicit_vs_explicit_roc} (a) only utilizes the geometric feature like Table \ref{tab:implicit_vs_explicit_geo_only}, whereas Fig. \ref{fig:implicit_vs_explicit_roc} (b) utilizes both geometric and texture feature like Table \ref{tab:implicit_vs_explicit_geo_tex}.
			As Fig. \ref{fig:implicit_vs_explicit_roc} (a) shows, our implicit methods can extract much better 3D representation even given sparse views compared with explicit methods with dense views.
			The EER of our FingerNeRF is raised from 22.60\% to 15.60\% by 7\% when the extra texture dimension is involved.
			From Fig. \ref{fig:implicit_vs_explicit_roc} (b), we can find out that although our FingerNeRF achieves slightly better EER compared with others, the performance gap between our implicit method and explicit methods are reduced significantly.
			The reason is that the information source of our FingerNeRF only cover the texture information of 3 views in total, whereas the explicit methods merge the texture information from all available views.
			
			\begin{table*}[t]
				\centering
				\caption{Ablation experiments of different rendered modalities on SCUT-Finger-3D dataset.}
				\resizebox{0.8\linewidth}{!}{
					\begin{tabular}{cc|cccccc}
						\hline
						\rowcolor{gray!40}
						\textbf{Render Image} & \textbf{Render Depth} & \textbf{EER} $^\downarrow$       & \textbf{T-F=0.01} $^\uparrow$  & \textbf{T-F=0.001} $^\uparrow$ & \textbf{mAP} $^\uparrow$       & \textbf{Rank 1} $^\uparrow$    & \textbf{Rank 5} $^\uparrow$ \\ \hline
						$\checkmark$ & $\times$     & 22.40\% & 30.00\% & 23.40\% & 48.78\% & 36.44\% & 58.37\%   \\
						$\times$     & $\checkmark$ & 22.60\% & 28.60\% & 25.00\% & 45.98\% & 36.30\% & 58.67\%   \\
						$\checkmark$ & $\checkmark$ & \textbf{15.60\%} & \textbf{42.00\%} & \textbf{29.00\%} & \textbf{53.97\%} & \textbf{41.93\%} & \textbf{68.00\%}   \\ \hline      
					\end{tabular}
				}
				\label{tab:ablation_render_modality}
				\vspace{-0.4cm}
			\end{table*}
			
			\begin{table*}[t]
				\centering
				\vspace{-0.4cm}
				\caption{Ablation experiments of different number of rendered views on SCUT-Finger-3D dataset.}
				\resizebox{0.8\linewidth}{!}{
					\begin{tabular}{cc|cccccc}
						\rowcolor{gray!40}
						\hline 
						\textbf{Input Views}   & \textbf{Rendered Views} & \textbf{EER} $^\downarrow$       & \textbf{T-F=0.01} $^\uparrow$  & \textbf{T-F=0.001} $^\uparrow$ & \textbf{mAP} $^\uparrow$       & \textbf{Rank 1} $^\uparrow$    & \textbf{Rank 5} $^\uparrow$ \\ \hline 
						3                               & 8              & 21.20\% & 35.00\% & 20.20\% & 53.14\% & 41.48\% & 65.78\%    \\
						3                               & 16             & 20.00\% & 41.60\% & 27.40\% & 57.20\% & 43.26\% & 66.37\%    \\
						3                               & 20             & \textbf{15.60\%} & \textbf{42.00\%} & \textbf{29.00\%} & \textbf{53.97\%} & \textbf{41.93\%} & \textbf{68.00\%}    \\ \hline
				\end{tabular}}
				\label{tab:ablation_render_views}
				\vspace{-0.4cm}
			\end{table*}

			\subsubsection{Comparison on SCUT-FingerVein-3D}
			\label{experiment:implicit_vs_explicit:scut_fingervein_3d}
			
			In this section, we provide the qualitative results of finger trait recognition in Table \ref{tab:implicit_vs_explicit_geo_only_fv} and \ref{tab:implicit_vs_explicit_geo_tex_fv}.
			Both of the verification and identification tasks are evaluated and the results of different methods are presented in the tables.
			Table \ref{tab:implicit_vs_explicit_geo_only_fv} compares the performance only using the 3D geometric information, and Table \ref{tab:implicit_vs_explicit_geo_tex_fv} compares the results with both 3D geometric and texture information.
			Moreover, we also visualize the DET curves of aforementioned experiments in Fig. \ref{fig:implicit_vs_explicit_roc_fv}.
			
			In Table \ref{tab:implicit_vs_explicit_geo_only_fv}, we can find that our FingerNeRF outperforms other explicit methods significantly on all evaluation metrics.
			In the verification task, the EER of our method is 25.28\% which is much better compared with the EERs of explicit methods ranging from 37\% to 40\%.
			In the identification task, the mAP of our method is 36.91\%, achieving an improvement of 9.67\% compared with explicit methods.

			In Table \ref{tab:implicit_vs_explicit_geo_tex_fv}, significant improvement can be witnessed in all evaluation metrics as well.
			The EER of our FingerNeRF is 16.98\%, outperforming explicit 3D methods by 13.37\%.
			For identification metrics, the mAP of our methods is 42.38\% which is higher than the best explicit 3D methods by 13.23\%.
			
			In Fig. \ref{fig:implicit_vs_explicit_roc_fv}, the DET curves of aforementioned experiments are presented.
			Fig. \ref{fig:implicit_vs_explicit_roc_fv}(a) only utilizes 3D geometric feature like Table \ref{tab:implicit_vs_explicit_geo_only_fv}, and Fig. \ref{fig:implicit_vs_explicit_roc_fv}(b) adopts both the geometric and texture feature like Table In Table \ref{tab:implicit_vs_explicit_geo_tex_fv}.
			As these Fig.s show, the EER of our FingerNeRF shows superior and robust performance compared with these explicit methods.

			\subsubsection{Ablation Study}
			\label{experiment:implicit_vs_explicit:ablation}

			\noindent\textbf{Effectiveness of different modalities:} 
			To evaluate the performance of different modalities rendered by our FingerNeRF, we provide the ablation results in Table \ref{tab:ablation_render_modality}.
			Only using single modality, the rendered images can provide slightly better performance than the rendered depth maps.
			For example, The EER of former modality is 22.40\%, and the one of latter is 22.60\%.
			The mAP of the former modality is 48.78\%, and the one of latter is 45.98\%.
			When both modalities are used, significant improvement can be witnessed in the last row of the table.

			\noindent\textbf{Effectiveness of different number of rendered views:} 
			As mentioned in Section \ref{experiment:implicit_vs_explicit:setting}, we render $V$ unseen views and feed the multi-view maps to multi-view convolution network for feature extraction.
			In Table \ref{tab:ablation_render_views}, we compare the results of verification and identification task under different number of rendered views.
			As the table shows, the performance improves along with the number of rendered views from 21.20\% ($V=8$) to 15.60\% ($V=16$).

			\subsection{Generalizable Neural Rendering}
			\label{experiment:generalizable_rendering}
			
			\subsubsection{Experiment Settings}
			\label{experiment:generalizable_rendering:settings}
			
			In this section, we aim to evaluate the rendering quality of the proposed implicit 3D reconstruction method.
			Since the finger trait recognition task is usually an open-set task in few-shot settings, the NeRF based methods are required to be generalizable towards sparse inputs.
			Instead of using the standard NeRF based methods which requires per-scene training and dense multi-view inputs, we select several generalizable NeRF based methods for comparison, including: PixelNeRF \cite{yu2021pixelnerf}, IBRNet \cite{wang2021ibrnet}, and MVSNeRF \cite{chen2021mvsnerf}.
			We use the released code of PixelNeRF, IBRNet and MVSNeRF, and train the models on our processed dataset. 
			We compare all these methods on 3 datasets with different modality of finger traits: SCUT-Finger-3D (Section \ref{experiment:generalizable_rendering:scut_finger_3d}), SCUT-FingerVein-3D (Section \ref{experiment:generalizable_rendering:scut_fingervein_3d}), and UNSW-3D (Section \ref{experiment:generalizable_rendering:unsw_3d}).
			The dataset split of train, valid, and test set is the same as Section \ref{experiment:implicit_vs_explicit}.
			In the training, validation, and testing phase, the same input views of 3 images are used. 
			Furthermore, we also provide the qualitative comparison with existing generalizable NeRF in Section \ref{experiment:generalizable_rendering:qual}.
			
			\subsubsection{Generalizable NeRF on SCUT-Finger-3D}
			\label{experiment:generalizable_rendering:scut_finger_3d}
			
			In this section, we compare the rendering quality of different methods under the generalizable setting on SCUT-Finger-3D dataset.
			In Table \ref{tab:gen_nvs_scut_finger_3d}, we provide the qualitative results with only 3 views as input.
			The metrics of PSNR, SSIM, and LPIPS are used to evaluate the rendering quality.
			The generalizable NeRF-based methods are firstly trained on the training set, and further evaluated on the unseen test set.
			From the table, we can find that the 3 methods can all achieve reasonable PSNR, SSIM, and LPIPS on the validation set and test set.
			As the table shows, the rendering results of our proposed FingerNeRF outperforms PixelNeRF, IBRNet, and MVSNeRF with the same inputs on all metrics.
			
			\begin{table}[t]
				\centering
				\footnotesize
				\vspace{-0.55cm}
				\caption{Experimental results of generalizable novel view synthesis on SCUT-Finger-3D dataset.}
				\label{tab:gen_nvs_scut_finger_3d}
				\setlength{\tabcolsep}{0.3mm}{
					\begin{tabular}{c|ccc}
						\hline
						\rowcolor{gray!40} 
						\textbf{Methods}        & \textbf{Test PSNR} $^\uparrow$ & \textbf{Test SSIM} $^\uparrow$ & \textbf{Test LPIPS} $^\downarrow$ \\ \hline
						PixelNeRF \cite{yu2021pixelnerf}               & 14.27 & 0.114 & 0.681     \\
						IBRNet \cite{wang2021ibrnet}                  & 27.37 & 0.540 & 0.271     \\
						MVSNeRF \cite{chen2021mvsnerf}                 & 29.09 & 0.567 & 0.193     \\ \hline
						FingerNeRF(Ours)         & \textbf{32.27} & \textbf{0.755} & \textbf{0.164}     \\ \hline
					\end{tabular}
				}
				\vspace{-0.6cm}
			\end{table}
			
			\begin{table}[t]
				\centering
				\footnotesize
				\vspace{-0.55cm}
				\caption{Experimental results of generalizable novel view synthesis on SCUT-FingerVein-3D dataset.}
				\label{tab:gen_nvs_scut_fingervein_3d}
				\setlength{\tabcolsep}{0.3mm}{
					\begin{tabular}{c|ccc}
						\hline
						\rowcolor{gray!40}
						\textbf{Methods}         & \textbf{Test PSNR} $^\uparrow$ & \textbf{Test SSIM} $^\uparrow$ & \textbf{Test LPIPS} $^\downarrow$  \\ \hline
						PixelNeRF \cite{yu2021pixelnerf}               & 16.14 & 0.294 & 0.612     \\
						IBRNet \cite{wang2021ibrnet}                  & 21.99 & 0.471 & 0.355     \\
						MVSNeRF \cite{chen2021mvsnerf}              & 24.77 & 0.692 & 0.244 \\ \hline
						FingerNeRF (Ours)                           & \textbf{26.67} & \textbf{0.819} & \textbf{0.235} \\ \hline
					\end{tabular}
				}
				\vspace{-0.6cm}
			\end{table}
			
			\begin{table}[t]
				\centering
				\footnotesize
				\vspace{-0.55cm}
				\caption{Experimental results of generalizable novel view synthesis on UNSW-3D dataset.}
				\label{tab:gen_nvs_unsw_3d}
				\setlength{\tabcolsep}{0.3mm}{
					\begin{tabular}{c|ccc}
						\hline
						\rowcolor{gray!40}
						\textbf{Methods}         & \textbf{Test PSNR} $^\uparrow$ & \textbf{Test SSIM} $^\uparrow$ & \textbf{Test LPIPS} $^\downarrow$  \\ \hline
						PixelNeRF \cite{yu2021pixelnerf}               & 16.61 & 0.359 & 0.523    \\
						IBRNet \cite{wang2021ibrnet}                  & 23.09 & 0.649 & 0.175    \\
						MVSNeRF \cite{chen2021mvsnerf}                 & 26.97 & 0.702 & 0.102    \\ \hline
						FingerNeRF(Ours)         & \textbf{30.97} & \textbf{0.801} & \textbf{0.064}    \\ \hline
					\end{tabular}
				}
				\vspace{-0.4cm}
			\end{table}

			\begin{table}[t]
				\centering
				\footnotesize
				\vspace{-0.55cm}
				\caption{Ablation experiments of the effectiveness of different components in FingerNeRF on SCUT-Finger-3D dataset.}
				\label{tab:gen_nvs_ablation_scut_finger_3d}
				\setlength{\tabcolsep}{0.3mm}{
					\begin{tabular}{cccc|ccc}
						\hline 
						\rowcolor{gray!40}
						\textbf{MVS}  & \textbf{Dep} & \textbf{Tra} & \textbf{Tran}  & \textbf{Test PSNR} $^\uparrow$ & \textbf{Test SSIM} $^\uparrow$ & \textbf{Test LPIPS} $^\downarrow$ \\ \hline 
						$\checkmark$     & $\times$ & $\times$  & $\times$              & 29.09 & 0.567 & 0.193           \\
						$\checkmark$     & $\checkmark$ & $\times$ & $\times$            & 29.39 & 0.618 & 0.189          \\  
						$\checkmark$     & $\checkmark$ & $\checkmark$ & $\times$        & 31.24 & 0.747 & 0.177          \\ 
						$\checkmark$     & $\checkmark$ & $\checkmark$ & $\checkmark$    & \textbf{32.27} & \textbf{0.755} & \textbf{0.164}          \\ \hline
				\end{tabular}}
				\vspace{-0.6cm}
			\end{table}
			
			\begin{table}[t]
				\centering
				\footnotesize
				\vspace{-0.55cm}
				\caption{Ablation experiments of the effectiveness of different components in FingerNeRF on SCUT-FingerVein-3D dataset.}
				\label{tab:gen_nvs_ablation_scut_fingervein_3d}
				\setlength{\tabcolsep}{0.3mm}{
					\begin{tabular}{cccc|ccc}
						\hline 
						\rowcolor{gray!40}
						\textbf{MVS}  & \textbf{Dep} & \textbf{Tra} & \textbf{Tran}  & \textbf{Test PSNR} $^\uparrow$ & \textbf{Test SSIM} $^\uparrow$ & \textbf{Test LPIPS} $^\downarrow$ \\ \hline 
						$\checkmark$     & $\times$ & $\times$  & $\times$               & 24.77 & 0.692 & 0.244 \\
						$\checkmark$     & $\checkmark$ & $\times$ & $\times$            & 25.07 & 0.752 & 0.241 \\  
						$\checkmark$     & $\checkmark$ & $\checkmark$ & $\times$        & 26.61 & 0.818 & 0.239 \\ 
						$\checkmark$     & $\checkmark$ & $\checkmark$ & $\checkmark$    & \textbf{26.67} & \textbf{0.819} & \textbf{0.235} \\ \hline
				\end{tabular}}
				\vspace{-0.6cm}
			\end{table}
			
			\begin{table}[t]
				\centering
				\footnotesize
				\vspace{-0.55cm}
				\caption{Ablation experiments of the effectiveness of different components in FingerNeRF on UNSW-3D dataset.}
				\label{tab:gen_nvs_ablation_unsw_3d}
				\setlength{\tabcolsep}{0.3mm}{
					\begin{tabular}{cccc|ccc}
						\hline 
						\rowcolor{gray!40}
						\textbf{MVS}  & \textbf{Dep} & \textbf{Tra} & \textbf{Tran}  & \textbf{Test PSNR} $^\uparrow$ & \textbf{Test SSIM} $^\uparrow$ & \textbf{Test LPIPS} $^\downarrow$ \\ \hline 
						$\checkmark$     & $\times$ & $\times$  & $\times$               & 29.09 & 0.567 & 0.193 \\
						$\checkmark$     & $\checkmark$ & $\times$ & $\times$            & 28.49 & 0.763 & 0.088 \\  
						$\checkmark$     & $\checkmark$ & $\checkmark$ & $\times$        & 29.28 & 0.731 & 0.078 \\ 
						$\checkmark$     & $\checkmark$ & $\checkmark$ & $\checkmark$    & \textbf{30.97} & \textbf{0.801} & \textbf{0.064} \\ \hline
				\end{tabular}}
				\vspace{-0.4cm}
			\end{table}
			
			\subsubsection{Generalizable NeRF on SCUT-Fingervein-3D}
			\label{experiment:generalizable_rendering:scut_fingervein_3d}
			
			In this section, we conduct generalizable neural rendering on another modality of finger vein images on SCUT-FingerVein dataset.
			In Table \ref{tab:gen_nvs_scut_fingervein_3d}, the experimental results with only finger vein images on 3 views as input are provided.
			The metrics of PSNR, SSIM, and LPIPS are used to evaluate the rendering quality.
			From the table, it can be found that the generalizable NeRF-based methods can still provide reasonable rendering results on the modality of finger vein images.
			Our FingerNeRF outperforms other methods on all 3 evaluation metrics of PSNR, SSIM, and LPIPS.
			Another interesting issue can be found that the PSNR of finger vein images (about 26) in Table \ref{tab:gen_nvs_scut_fingervein_3d} are lower than the PSNR of finger images (about 32) in Table \ref{tab:gen_nvs_scut_finger_3d}.
			The reason is that the raw finger images in SCUT-Finger-3D dataset has abundant fingerprint texture in local regions, but the finger vein images contains less effective textures and more noises because of the transmission imaging mechanism to capture vein textures.

			\begin{figure*}[t]%
				\centering
				\includegraphics[width=0.9\textwidth]{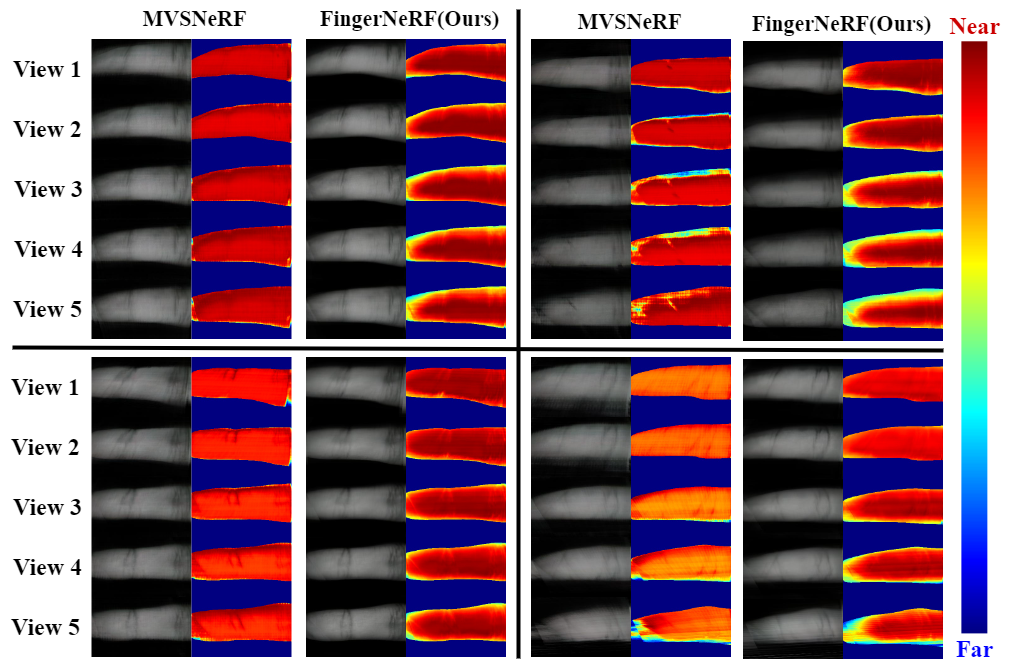}
				\caption{Qualitative comparison with state-of-the-art generalizable NeRF(MVSNeRF \cite{chen2021mvsnerf}).}
				\label{fig:comparison_with_mvsnerf}
				\vspace{-0.4cm}
			\end{figure*}
			
			\begin{table*}[t]
				\centering
				\caption{Comparison with Mulit-view 3D shape recognition methods on SCUT-Finger-3D dataset.}
				\label{tab:multi_view_comp_scut_finger_3d}
				\resizebox{0.9\linewidth}{!}{
					\begin{tabular}{l|ccccccc}
						\hline
						\rowcolor{gray!40}
						\textbf{Method}                       & \textbf{EER} $^\downarrow$       & \textbf{T-F=0.01} $^\uparrow$  & \textbf{T-F=0.001} $^\uparrow$ & \textbf{mAP} $^\uparrow$       & \textbf{Rank 1} $^\uparrow$    & \textbf{Rank 5} $^\uparrow$  \\ 
						\hline
						MVCNN \cite{su2015multi} + ResNet-18 \cite{he2016deep}            & 8.35\%    & 60.08\%   & 33.10\%   & 67.20\%   & 65.06\%   & 92.66\% \\
						MVCNN \cite{su2015multi} + ResNet-50 \cite{he2016deep}            & 8.75\%    & 63.70\%   & 40.02\%   & 75.57\%   & 71.22\%   & 92.44\% \\
						MVCNN \cite{su2015multi} + ShuffleNet v2\_10 \cite{ma2018shufflenet}   & 16.98\%   & 38.32\%   & 28.22\%   & 53.83\%   & 47.21\%   & 72.82\% \\
						MVCNN \cite{su2015multi} + MobileNet\_V2 \cite{sandler2018mobilenetv2}       & 5.08\%    & 80.92\%   & 44.70\%   & 84.07\%   & 78.72\%   & 95.96\% \\
						MVCNN \cite{su2015multi} + MobileNet\_V3\_large \cite{howard2019searching} & 6.01\%    & 72.68\%   & 49.46\%   & 80.75\%   & 77.69\%   & 94.55\% \\
						MVCNN \cite{su2015multi} + EfficientNet-B1 \cite{tan2019efficientnet}     & 5.44\%    & 77.30\%   & 51.32\%   & 80.00\%   & 80.06\%   & 95.87\% \\
						MVCNN \cite{su2015multi} + mnasnet\_10 \cite{tan2019mnasnet}         & 12.20\%   & 50.46\%   & 34.29\%   & 64.30\%   & 62.15\%   & 89.39\% \\
						MVCNN \cite{su2015multi} + regnet\_x\_800mf \cite{radosavovic2020designing}     & 7.66\%    & 67.60\%   & 39.12\%   & 72.00\%   & 68.72\%   & 93.75\% \\ 
						RotNet \cite{kanezaki2018rotationnet} + ResNet 18 \cite{he2016deep}           & 5.67\%    & 77.20\%   & 45.60\%   & 79.78\%   & 75.71\%   & 96.99\% \\
						RotNet \cite{kanezaki2018rotationnet} + ResNet 50 \cite{he2016deep}           & 5.65\% & 75.88\% & 41.40\% & 81.01\% & 76.89\% & 96.03\% \\
						RotNet \cite{kanezaki2018rotationnet} + ShuffleNet v2\_10 \cite{ma2018shufflenet}   & 17.56\%	 & 34.08\%	 & 13.70\%	 & 42.41\%	 & 40.35\%	 & 70.80\%   \\
						RotNet \cite{kanezaki2018rotationnet} + MobileNet\_V2 \cite{sandler2018mobilenetv2}       & 5.00\%	 & 81.30\%	 & 56.94\%	 & 85.87\%	 & 82.15\%	 & 96.03\%   \\
						RotNet \cite{kanezaki2018rotationnet} + MobileNet\_V3\_large \cite{howard2019searching} & 5.69\%	 & 78.22\%	 & 55.82\%	 & 82.70\%	 & 80.06\%	 & 95.54\%   \\
						RotNet \cite{kanezaki2018rotationnet} + EfficientNet-B1 \cite{tan2019efficientnet}     & 5.07\%	 & 79.28\%	 & 55.26\%	 & 89.26\%	 & 84.39\%	 & 96.24\%   \\
						RotNet \cite{kanezaki2018rotationnet} + mnasnet\_10 \cite{tan2019mnasnet}         & 26.81\%	 & 34.04\%	 & 28.00\%	 & 47.91\%	 & 38.94\%	 & 54.49\%   \\
						MHBN \cite{yu2018multi}                         & 28.25\%	 & 29.74\%	 & 23.68\%	 & 33.66\%	 & 34.65\%	 & 50.03\%   \\
						MHBN \cite{yu2018multi} + ResNet 18  \cite{he2016deep}            & 18.76\%	 & 30.04\%	 & 16.96\%	 & 38.65\%	 & 36.86\%	 & 64.68\%   \\
						MHBN \cite{yu2018multi} + ResNet 50  \cite{he2016deep}            & 18.29\%	 & 33.08\%	 & 20.16\%	 & 37.37\%	 & 35.19\%	 & 66.28\%   \\
						MHBN \cite{yu2018multi} + ShuffleNet v2\_10 \cite{ma2018shufflenet}     & 21.44\%	 & 20.58\%	 & 5.62\%	 & 38.94\%	 & 32.88\%	 & 65.03\%   \\
						MHBN \cite{yu2018multi} + MobileNet\_V2 \cite{sandler2018mobilenetv2}         & 20.01\%	 & 33.52\%	 & 13.14\%	 & 48.18\%	 & 38.65\%	 & 66.15\%   \\
						MHBN \cite{yu2018multi} + MobileNet\_V3\_large  \cite{howard2019searching} & 21.10\%	 & 27.06\%	 & 17.64\%	 & 50.71\%	 & 41.73\%	 & 62.69\%   \\
						MHBN \cite{yu2018multi} + EfficientNet-B1
						\cite{tan2019efficientnet}      & 19.23\%	 & 31.30\%	 & 16.00\%	 & 40.05\%	 & 40.26\%	 & 65.26\%   \\
						\hline
						Yang. \cite{yang2021lfmb}                              & 8.35\%   & 60.08\%    & 33.10\%   & 67.20\%   & 65.06\%   & 92.66\% \\
						Lin. \cite{lin2018contactless} + ResNet 18 \cite{he2016deep}            & 5.61\%	& 77.10\%	 & 52.60\%	 & 78.49\%   & 75.93\%   & 96.47\% \\  
						Lin. \cite{lin2018contactless} + ResNet 50 \cite{he2016deep}            & 6.14\%	& 69.44\%	 & 48.62\%	 & 75.25\%   & 73.17\%   & 95.87\% \\ \hline
						FingerNeRF (Ours) & \textbf{4.37\%} & \textbf{81.54\%} & \textbf{60.62\%} & \textbf{88.84\%} & \textbf{83.88\%} & \textbf{97.05\%} \\
						\hline
					\end{tabular}
				}
				\vspace{-0.4cm}
			\end{table*}

			\begin{table*}[t]
				\centering
				\caption{Comparison with Mulit-view 3D shape recognition methods on SCUT-FingerVein-3D dataset.}
				\label{tab:multi_view_comp_scut_fingervein_3d}
				\resizebox{0.9\linewidth}{!}{
					\begin{tabular}{l|ccccccc}
						\hline
						\rowcolor{gray!40}
						\textbf{Method}                       & \textbf{EER} $^\downarrow$       & \textbf{T-F=0.01} $^\uparrow$  & \textbf{T-F=0.001} $^\uparrow$ & \textbf{mAP} $^\uparrow$       & \textbf{Rank 1} $^\uparrow$    & \textbf{Rank 5} $^\uparrow$  \\ 
						\hline
						MVCNN \cite{su2015multi} + ResNet-18 \cite{he2016deep}            & 12.52\% & 48.25\% & 32.92\% & 50.73\% & 50.00\% & 81.22\% \\
						MVCNN \cite{su2015multi} + ResNet-50 \cite{he2016deep}            & 11.75\% & 43.30\% & 27.92\% & 57.30\% & 57.54\% & 85.30\% \\ 
						MVCNN \cite{su2015multi} + ShuffleNet v2\_10 \cite{ma2018shufflenet}   & 17.48\% & 29.09\% & 17.74\% & 38.14\% & 37.39\% & 66.03\% \\
						MVCNN \cite{su2015multi} + MobileNet\_V2 \cite{sandler2018mobilenetv2}      & 10.75\% & 55.45\% & 35.26\% & 61.59\% & 62.59\% & 88.23\% \\
						MVCNN \cite{su2015multi} + MobileNet\_V3\_large \cite{howard2019searching} & 12.63\% & 50.85\% & 31.92\% & 55.39\% & 55.87\% & 80.85\% \\
						MVCNN \cite{su2015multi} + EfficientNet-B1 \cite{tan2019efficientnet}     &  9.06\% & 62.85\% & 28.65\% & 63.99\% & 63.63\% & 89.75\% \\
						MVCNN \cite{su2015multi} + mnasnet\_10 \cite{tan2019mnasnet}         & 12.42\% & 46.38\% & 27.04\% & 52.49\% & 54.33\% & 54.33\% \\
						MVCNN \cite{su2015multi} + regnet\_x\_800mf \cite{radosavovic2020designing}     & 11.27\% & 42.96\% & 27.17\% & 58.99\% & 52.54\% & 84.18\% \\ 
						RotNet \cite{kanezaki2018rotationnet} + ResNet 18 \cite{he2016deep}           & 10.62\% & 51.42\% & 33.72\% & 61.27\% & 61.27\% & 87.24\% \\
						RotNet \cite{kanezaki2018rotationnet} + ResNet 50 \cite{he2016deep}           & 10.42\% & 46.64\% & 28.21\% & 58.35\% & 53.81\% & 85.03\% \\
						RotNet \cite{kanezaki2018rotationnet} + ShuffleNet v2\_10 \cite{ma2018shufflenet}   & 14.23\% & 21.64\% &  0.00\% & 32.37\% & 28.37\% & 70.30\%  \\
						RotNet \cite{kanezaki2018rotationnet} + MobileNet\_V2 \cite{sandler2018mobilenetv2}     & 9.23\% & 52.17\% & 32.30\% & 63.81\% & 63.81\% & 89.05\% \\
						RotNet \cite{kanezaki2018rotationnet} + MobileNet\_V3\_large \cite{howard2019searching} & 8.90\% & 59.49\% & 35.77\% & 62.86\% & 61.92\% & 91.32\% \\
						RotNet \cite{kanezaki2018rotationnet} + EfficientNet-B1 \cite{tan2019efficientnet}     &  8.76\% & 60.98\% & 34.58\% & 66.63\% & 65.91\% & 89.96\%   \\
						RotNet \cite{kanezaki2018rotationnet} + mnasnet\_10 \cite{tan2019mnasnet}         & 36.59\% & 19.65\% & 12.25\% & 23.01\% & 20.19\% & 39.74\%   \\
						MHBN \cite{yu2018multi}                         & 36.59\% & 19.65\% & 12.25\% & 23.01\% & 20.19\% & 39.74\%   \\
						MHBN \cite{yu2018multi} + ResNet 18  \cite{he2016deep}            & 20.16\% & 22.30\% &  7.32\% & 34.96\% & 32.73\% & 62.46\%  \\
						MHBN \cite{yu2018multi} + ResNet 50  \cite{he2016deep}            & 18.86\% & 21.00\% & 15.02\% & 30.46\% & 31.76\% & 63.01\%   \\
						MHBN \cite{yu2018multi} + ShuffleNet v2\_10 \cite{ma2018shufflenet}     & 18.77\% & 12.66\% &  3.91\% & 29.84\% & 28.68\% & 60.01\%   \\
						MHBN \cite{yu2018multi} + MobileNet\_V2 \cite{sandler2018mobilenetv2}      & 38.75\% & 4.04\% & 4.20\% & 4.20\% & 5.93\% & 22.08\%   \\
						MHBN \cite{yu2018multi} + MobileNet\_V3\_large  \cite{howard2019searching} & 34.00\% & 3.23\% & 4.10\% & 8.53\% & 9.17\% & 19.20\%   \\
						MHBN \cite{yu2018multi} + EfficientNet-B1 \cite{tan2019efficientnet}      & 35.99\% &  6.13\% &  1.51\% &  8.77\% & 10.28\% & 29.70\%   \\
						\hline
						Yang. \cite{yang2021lfmb}                                         & 11.91\% & 44.13\% & 25.38\% & 46.57\% & 45.64\% & 83.39\% \\
						Lin. \cite{lin2018contactless} + ResNet 18 \cite{he2016deep}      & 10.38\% & 53.21\% & 34.30\% & 58.23\% & 56.11\% & 85.00\% \\ 
						Lin. \cite{lin2018contactless} + ResNet 50 \cite{he2016deep}      & 11.28\% & 49.38\% & 32.38\% & 49.38\% & 55.44\% & 85.72\% \\ \hline
						FingerNeRF (Ours)                                                     & \textbf{8.12\%} & \textbf{63.19\%} & \textbf{36.68\%} & \textbf{68.35\%} & \textbf{66.00\%} & \textbf{91.89\%} \\
						\hline
					\end{tabular}
				}
				\vspace{-0.4cm}
			\end{table*}
			
			\begin{table*}[t]
				\centering
				\caption{Comparison with Mulit-view 3D shape recognition methods on UNSW-3D dataset.}
				\label{tab:multi_view_comp_unsw_3d}
				\resizebox{0.9\linewidth}{!}{
					\begin{tabular}{l|ccccccc}
						\hline
						\rowcolor{gray!40}
						\textbf{Method}                       & \textbf{EER} $^\downarrow$       & \textbf{T-F=0.01} $^\uparrow$  & \textbf{T-F=0.001} $^\uparrow$ & \textbf{mAP} $^\uparrow$       & \textbf{Rank 1} $^\uparrow$    & \textbf{Rank 5} $^\uparrow$  \\ 
						\hline
						MVCNN \cite{su2015multi} + ResNet-18 \cite{he2016deep}            & 17.48\%	& 53.67\%	& 32.52\%	& 15.64\%	& 14.65\%	& 25.83\% \\
						MVCNN \cite{su2015multi} + ResNet-50 \cite{he2016deep}            & 16.82\%	& 55.68\%	& 39.87\%	& 16.13\%	& 15.98\%	& 25.97\% \\
						MVCNN \cite{su2015multi} + ShuffleNet v2\_10 \cite{ma2018shufflenet}   & 13.70\%	& 66.15\%	& 48.78\%	& 18.98\%	& 17.98\%	& 30.36\% \\
						MVCNN \cite{su2015multi} + MobileNet\_V2 \cite{sandler2018mobilenetv2}       & 3.56\%	& 92.65\%	& 81.96\%	& 24.37\%	& 24.37\%	& 37.28\% \\
						MVCNN \cite{su2015multi} + MobileNet\_V3\_large \cite{howard2019searching} & 7.57\%	& 78.17\%	& 43.88\%	& 20.77\%	& 19.84\%	& 33.29\% \\
						MVCNN \cite{su2015multi} + EfficientNet-B1 \cite{tan2019efficientnet}     & 5.57\%	& 86.41\%	& 76.17\%	& 22.33\%	& 24.50\%	& 38.75\% \\
						MVCNN \cite{su2015multi} + mnasnet\_10 \cite{tan2019mnasnet}         & 8.57\%	& 83.07\%	& 69.27\%	& 21.21\%	& 23.97\%	& 35.29\% \\
						MVCNN \cite{su2015multi} + regnet\_x\_800mf \cite{radosavovic2020designing}     & 12.58\%	& 71.49\%	& 58.80\%	& 16.16\%	& 18.38\%	& 30.36\% \\ 
						RotNet \cite{kanezaki2018rotationnet} + ResNet 18 \cite{he2016deep}           & 4.01\%	& 89.53\%	& 79.51\%	& 22.84\%	& 23.57\%	& 35.95\% \\
						RotNet \cite{kanezaki2018rotationnet} + ResNet 50 \cite{he2016deep}           & 3.56\%	& 93.32\%	& 81.74\%	& 21.92\%	& 24.37\%	& 37.42\% \\
						RotNet \cite{kanezaki2018rotationnet} + ShuffleNet v2\_10 \cite{ma2018shufflenet}   & 12.69\%	& 69.71\%	& 62.36\%	& 17.56\%	& 19.97\%	& 32.22\% \\
						RotNet \cite{kanezaki2018rotationnet} + MobileNet\_V2 \cite{sandler2018mobilenetv2}       & 3.56\%	& 92.65\%	& 81.96\%	& 24.37\%	& 24.37\%	& 37.28\% \\
						RotNet \cite{kanezaki2018rotationnet} + MobileNet\_V3\_large \cite{howard2019searching} & 4.23\%	& 92.65\%	& 81.74\%	& 23.20\%	& 23.70\%	& 35.55\%  \\
						RotNet \cite{kanezaki2018rotationnet} + EfficientNet-B1 \cite{tan2019efficientnet}     & 3.79\% & 94.21\% & 78.62\% & 23.89\% & 24.50\% & 39.41\% \\
						RotNet \cite{kanezaki2018rotationnet} + mnasnet\_10 \cite{tan2019mnasnet}         & 8.69\%	& 85.52\%	& 73.72\%	& 20.45\%	& 21.44\%	& 33.95\%   \\
						MHBN \cite{yu2018multi}                        & 27.39\%	& 10.02\%	& 0.45\%	& 5.85\%	& 4.79\%	& 10.92\%  \\
						MHBN \cite{yu2018multi} + ResNet 18  \cite{he2016deep}            & 24.94\%	& 34.52\%	& 25.84\%	& 5.91\%	& 5.86\%	& 14.38\%  \\
						MHBN \cite{yu2018multi} + ResNet 50  \cite{he2016deep}            & 29.96\%	& 16.70\%	& 1.11\%	& 5.18\%	& 5.19\%	& 12.25\%  \\
						MHBN \cite{yu2018multi} + ShuffleNet v2\_10 \cite{ma2018shufflenet}     & 21.60\%	& 27.39\%	& 1.34\%	& 5.34\%	& 4.93\%	& 14.38\% \\
						MHBN \cite{yu2018multi} + MobileNet\_V2 \cite{sandler2018mobilenetv2}         & 18.37\%	& 42.09\%	& 16.48\%	& 8.85\%	& 8.92\%	& 18.51\%   \\
						MHBN \cite{yu2018multi} + MobileNet\_V3\_large  \cite{howard2019searching} &  16.48\%	& 47.88\%	& 26.28\%	& 8.50\%	& 11.05\%	& 22.10\%  \\
						MHBN \cite{yu2018multi} + EfficientNet-B1 \cite{tan2019efficientnet}      & 21.94\%	& 31.40\%	& 5.79\%	& 8.27\%	& 7.32\%	& 15.98\%  \\
						\hline
						Yang. \cite{yang2021lfmb}                              & 16.37\%	& 41.20\%	& 26.95\%	& 12.47\%	& 10.65\%	& 19.84\% \\
						Lin. \cite{lin2018contactless} + ResNet 18 \cite{he2016deep}            & 5.46\%	& 85.52\%	& 60.13\%	& 18.74\%	& 20.77\%	& 31.42\% \\ 
						Lin. \cite{lin2018contactless} + ResNet 50 \cite{he2016deep}            & 4.01\%	& 92.43\%	& 86.19\%	& 23.77\%	& 23.44\%	& 34.89\% \\ \hline
						FingerNeRF (Ours) & \textbf{2.90\%} & \textbf{94.65\%} & \textbf{87.31\%} & \textbf{25.49\%} & \textbf{25.30\%} & \textbf{39.55\%} \\
						\hline
					\end{tabular}
				} 
				\vspace{-0.4cm}
			\end{table*}
			
			\subsubsection{Generalizable NeRF on UNSW-3D}
			\label{experiment:generalizable_rendering:unsw_3d}
			
			In this section, we aim to evaluate the generalization ability of the proposed methods towards existing 3D finger trait dataset, UNSW-3D.
			Different from SCUT-Finger-3D and SCUT-FingerVein-3D dataset, UNSW-3D only contains 3 view images rather the video sequence.
			Since the 3 view setting is not enough for training an entire generalizable NeRF, we firstly borrow the pretrained model from SCUT-Finger-3D and finetune the same model on the UNSW-3D dataset.
			In the finetuning phase, we randomly select 2 images from the given 3 views and the remaining one as the target view for rendering and loss computation.
			The qualitative results are provided in Table \ref{tab:gen_nvs_unsw_3d}.
			It can be found from the table that the proposed FingerNeRF outperforms other methods on all evaluation metrics.
			The experimental results prove the the proposed method can also be generalized to existing 3d finger trait datasets with multi-view configuration which may inspires more researches in the future.

			\subsubsection{Qualitative Comparison}
			\label{experiment:generalizable_rendering:qual}
			
			In Fig. \ref{fig:comparison_with_mvsnerf}, we provide qualitative comparison results between the proposed FingerNeRF and the representative method of generalizable NeRF, MVSNeRF \cite{chen2021mvsnerf}.
			The rendered images and depth maps on a continuous sequence of camera trajectories are shown in the figure.
			Due to the shape-radiance ambiguity, we can find that the rendered depth map of MVSNeRF tends to be incorrect.
			The depth values on the edge area and the center area of finger have no clear difference.
			It does not make sense, considering that the shape of finger is similar to an elliptic cylinder.
			Whereas our method can render both the reasonable depth maps and images as shown in the figure.

			\subsubsection{Ablation Study}
			\label{experiment:generalizable_rendering:ablation}
			
			In this part, we provide the ablation study of the effectiveness on different components in the proposed FingerNeRF.
			The qualitative results on 3 datasets of SCUT-Finger-3D, SCUT-FingerVein-3D, and UNSW-3D are respectively provided in Table \ref{tab:gen_nvs_ablation_scut_finger_3d}, \ref{tab:gen_nvs_ablation_scut_fingervein_3d}, and \ref{tab:gen_nvs_ablation_unsw_3d}.
			In the tables, \emph{MVS} represents the baseline NeRF based model, MVSNeRF, finetuned on the corresponding dataset.
			\emph{Dep} represents the depth distillation loss $L_{\text{dep}}$ in Eq. \ref{eq20}.
			\emph{Tra} means the trait guided neural rendering loss $L_{\text{tra}}$ in Eq. \ref{eq24}.
			\emph{Trans} is the proposed attention-based transformer module in the FingerNeRF.
			It can be found that the proposed components are beneficial in all 3 mentioned datasets, even across different modalities like finger vein and finger images.
			The involvement of depth supervision can insert extra depth priors into the learning framework of NeRF, and improve the rendering performance with a better representation of the 3D shape.
			As the tables show, the depth supervision term can improve all 3 metrics on all 3 datasets.
			For the proposed gradient loss, the constraint on local patches can mitigate the oversmoothing problem that may lose fine-grained information during rendering.
			As the improvement of PSNR/SSIM/LPIPS in the tables show, the proposed transformer-based enforcement module for feature extraction with finger trait priors can effectively involve the correspondence between finger traits like fingerprint and vein textures on epipolar lines among different views.
			
			\subsection{Multi-view Recognition}
			\label{experiment:multi_view}

			\subsubsection{Experiment Settings}
			\label{experiment:multi_view:settings}

			In this section, we aim to find out whether the proposed method can be used to improve the performance when compared with existing multi-view recognition methods.
			Different from Section \ref{experiment:implicit_vs_explicit} that only construct single multi-view pair by randomly selecting 3 views from the whole video sequence, we build multiple multi-view pairs by enumerating different combinations of available views in the video.
			In this way, more variation of multi-view pairs can be used to represent the subject and all available views of the captured videos are used for training.
			We still follow the same data split of 5/2/3 in train/valid/test set, like the experiments in Section \ref{experiment:implicit_vs_explicit}.
			For the compared methods, we utilize different multi-view recognition networks with different network backbones.
			The existing state-of-the-art multi-view recognition network structure includes: MVCNN \cite{su2015multi}, RotNet \cite{kanezaki2018rotationnet}, and MHBN \cite{yu2018multi}.
			Furthermore, multi-view recognition networks specifically designed for finger traits, like Yang. \cite{yang2021lfmb} for finger vein, and Lin. \cite{lin2018contactless} for fingerprint, are used for comparison as well.
			The network backbones includes ResNet-18/34/50 \cite{he2016deep}, ShuffleNet v2 \cite{ma2018shufflenet}, MobileNet\_V2 \cite{sandler2018mobilenetv2}, MobileNet\_V3 \cite{howard2019searching}, EfficientNet \cite{tan2019efficientnet}, mnasnet \cite{tan2019mnasnet}, and regnet \cite{radosavovic2020designing}.
			Considering the trade-off between efficiency and accuracy, we only utilize the 3 view images as input to the network like Lin. \cite{lin2018contactless}.
			For our proposed method, we input the given 3 view images to the FingerNeRF and render the depth maps on the corresponding 3 views.
			The utilized FingerNeRF is pretrained on the same dataset split of train/valid/test set in Section \ref{experiment:generalizable_rendering}.
			The depth maps are further concatenated with the original images on the same view, and fed to the multi-view recognition network.
			We utilize RotNet as the default multi-view recognition network architecture and EfficientNet as the backbone network.
			We conduct experiments on all 3 datasets, including SCUT-Finger-3D (Section \ref{experiment:multi_view:scut_finger_3d}), SCUT-FingerVein-3D (Section \ref{experiment:multi_view:scut_fingervein_3d}), and UNSW-3D (Section \ref{experiment:multi_view:unsw_3d}).

				\renewcommand{\thefigure}{\arabic{figure}}
				
				\renewcommand{\thetable}{\arabic{table}}

				\begin{table}[t]
					\centering
					\footnotesize
					\vspace{-0.55cm}
					\caption{Experimental results of generalizable novel view synthesis on SCUT-Finger-3D dataset.}
					\label{tab:gen_nvs_scut_finger_3d_appendix}
					\setlength{\tabcolsep}{0.3mm}{
						\begin{tabular}{c|ccc}
							\hline
							\rowcolor{gray!40}
							\textbf{Methods}         & \textbf{Val PSNR} $^\uparrow$ & \textbf{Val SSIM} $^\uparrow$ & \textbf{Val LPIPS} $^\downarrow$  \\ \hline
							PixelNeRF \cite{yu2021pixelnerf}               & 14.93 & 0.098 & 0.668     \\
							IBRNet \cite{wang2021ibrnet}                  & 24.93 & 0.498 & 0.256      \\
							MVSNeRF \cite{chen2021mvsnerf}                 & 27.57 & 0.499 & 0.213      \\ \hline
							FingerNeRF(Ours)         & \textbf{31.34} & \textbf{0.716} & \textbf{0.199}      \\ \hline
							\rowcolor{gray!40} 
							\textbf{Methods}        & \textbf{Test PSNR} $^\uparrow$ & \textbf{Test SSIM} $^\uparrow$ & \textbf{Test LPIPS} $^\downarrow$ \\ \hline
							PixelNeRF \cite{yu2021pixelnerf}               & 14.27 & 0.114 & 0.681     \\
							IBRNet \cite{wang2021ibrnet}                  & 27.37 & 0.540 & 0.271     \\
							MVSNeRF \cite{chen2021mvsnerf}                 & 29.09 & 0.567 & 0.193     \\ \hline
							FingerNeRF(Ours)         & \textbf{32.27} & \textbf{0.755} & \textbf{0.164}     \\ \hline
						\end{tabular}
					}
					\vspace{-0.6cm}
				\end{table}
				
				\begin{table}[t]
					\centering
					\footnotesize
					\vspace{-0.55cm}
					\caption{Experimental results of generalizable novel view synthesis on SCUT-FingerVein-3D dataset.}
					\label{tab:gen_nvs_scut_fingervein_3d_appendix}
					\setlength{\tabcolsep}{0.3mm}{
						\begin{tabular}{c|ccc}
							\hline
							\rowcolor{gray!40}
							\textbf{Methods}         & \textbf{Val PSNR} $^\uparrow$ & \textbf{Val SSIM} $^\uparrow$ & \textbf{Val LPIPS} $^\downarrow$   \\ \hline
							PixelNeRF \cite{yu2021pixelnerf}               & 12.32 & 0.203 & 0.728      \\
							IBRNet \cite{wang2021ibrnet}                  & 17.34 & 0.406 & 0.416      \\
							MVSNeRF \cite{chen2021mvsnerf}              & 22.69 & 0.668 & 0.239  \\ \hline
							FingerNeRF (Ours)                           & \textbf{24.39} & \textbf{0.809} & \textbf{0.227}  \\ \hline
							\rowcolor{gray!40}
							\textbf{Methods}         & \textbf{Test PSNR} $^\uparrow$ & \textbf{Test SSIM} $^\uparrow$ & \textbf{Test LPIPS} $^\downarrow$  \\ \hline
							PixelNeRF \cite{yu2021pixelnerf}               & 16.14 & 0.294 & 0.612     \\
							IBRNet \cite{wang2021ibrnet}                  & 21.99 & 0.471 & 0.355     \\
							MVSNeRF \cite{chen2021mvsnerf}              & 24.77 & 0.692 & 0.244 \\ \hline
							FingerNeRF (Ours)                           & \textbf{26.67} & \textbf{0.819} & \textbf{0.235} \\ \hline
						\end{tabular}
					}
					\vspace{-0.6cm}
				\end{table}
				
				\begin{table}[t]
					\centering
					\footnotesize
					\vspace{-0.55cm}
					\caption{Experimental results of generalizable novel view synthesis on UNSW-3D dataset.}
					\label{tab:gen_nvs_unsw_3d_appendix}
					\setlength{\tabcolsep}{0.3mm}{
						\begin{tabular}{c|ccc}
							\hline
							\rowcolor{gray!40}
							\textbf{Methods}         & \textbf{Val PSNR} $^\uparrow$ & \textbf{Val SSIM} $^\uparrow$ & \textbf{Val LPIPS} $^\downarrow$   \\ \hline
							PixelNeRF \cite{yu2021pixelnerf}               & 16.88 & 0.384 & 0.518     \\
							IBRNet \cite{wang2021ibrnet}                  & 23.32 & 0.639 & 0.174     \\
							MVSNeRF \cite{chen2021mvsnerf}                 & 27.42 & 0.707 & 0.093     \\ \hline
							FingerNeRF(Ours)         & \textbf{31.32} & \textbf{0.813} & \textbf{0.062}    \\ \hline
							\rowcolor{gray!40}
							\textbf{Methods}         & \textbf{Test PSNR} $^\uparrow$ & \textbf{Test SSIM} $^\uparrow$ & \textbf{Test LPIPS} $^\downarrow$  \\ \hline
							PixelNeRF \cite{yu2021pixelnerf}               & 16.61 & 0.359 & 0.523    \\
							IBRNet \cite{wang2021ibrnet}                  & 23.09 & 0.649 & 0.175    \\
							MVSNeRF \cite{chen2021mvsnerf}                 & 26.97 & 0.702 & 0.102    \\ \hline
							FingerNeRF(Ours)         & \textbf{30.97} & \textbf{0.801} & \textbf{0.064}    \\ \hline
						\end{tabular}
					}
					\vspace{-0.2cm}
				\end{table}

				\begin{table}[t]
					\centering
					\footnotesize
					\vspace{-0.55cm}
					\caption{Ablation experiments of the effectiveness of different components in FingerNeRF on SCUT-Finger-3D dataset.}
					\label{tab:gen_nvs_ablation_scut_finger_3d_appendix}
					\setlength{\tabcolsep}{0.3mm}{
						\begin{tabular}{cccc|ccc}
							\hline 
							\rowcolor{gray!40}
							\textbf{MVS}  & \textbf{Dep} & \textbf{Tra} & \textbf{Tran}  & \textbf{Val PSNR} $^\uparrow$ & \textbf{Val SSIM} $^\uparrow$ & \textbf{Val LPIPS} $^\downarrow$  \\ \hline 
							$\checkmark$     & $\times$ & $\times$  & $\times$               & 27.57 & 0.499 & 0.213            \\
							$\checkmark$     & $\checkmark$ & $\times$ & $\times$            & 29.10 & 0.534 & 0.213           \\  
							$\checkmark$     & $\checkmark$ & $\checkmark$ & $\times$        & 29.95 & 0.697 & 0.204           \\ 
							$\checkmark$     & $\checkmark$ & $\checkmark$ & $\checkmark$    & \textbf{31.34} & \textbf{0.716} & \textbf{0.199}           \\ \hline
							\rowcolor{gray!40}
							\textbf{MVS}  & \textbf{Dep} & \textbf{Tra} & \textbf{Tran}  & \textbf{Test PSNR} $^\uparrow$ & \textbf{Test SSIM} $^\uparrow$ & \textbf{Test LPIPS} $^\downarrow$ \\ \hline 
							$\checkmark$     & $\times$ & $\times$  & $\times$              & 29.09 & 0.567 & 0.193           \\
							$\checkmark$     & $\checkmark$ & $\times$ & $\times$            & 29.39 & 0.618 & 0.189          \\  
							$\checkmark$     & $\checkmark$ & $\checkmark$ & $\times$        & 31.24 & 0.747 & 0.177          \\ 
							$\checkmark$     & $\checkmark$ & $\checkmark$ & $\checkmark$    & \textbf{32.27} & \textbf{0.755} & \textbf{0.164}          \\ \hline
					\end{tabular}}
					\vspace{-0.6cm}
				\end{table}
				
				\begin{table}[t]
					\centering
					\footnotesize
					\vspace{-0.55cm}
					\caption{Ablation experiments of the effectiveness of different components in FingerNeRF on SCUT-FingerVein-3D dataset.}
					\label{tab:gen_nvs_ablation_scut_fingervein_3d_appendix}
					\setlength{\tabcolsep}{0.3mm}{
						\begin{tabular}{cccc|ccc}
							\hline 
							\rowcolor{gray!40}
							\textbf{MVS}  & \textbf{Dep} & \textbf{Tra} & \textbf{Tran}  & \textbf{Val PSNR} $^\uparrow$ & \textbf{Val SSIM} $^\uparrow$ & \textbf{Val LPIPS} $^\downarrow$  \\ \hline 
							$\checkmark$     & $\times$ & $\times$  & $\times$               & 22.69 & 0.668 & 0.239  \\
							$\checkmark$     & $\checkmark$ & $\times$ & $\times$            & 23.50 & 0.742 & 0.246  \\  
							$\checkmark$     & $\checkmark$ & $\checkmark$ & $\times$        & 24.28 & 0.805 & 0.231  \\ 
							$\checkmark$     & $\checkmark$ & $\checkmark$ & $\checkmark$    & \textbf{24.39} & \textbf{0.809} & \textbf{0.227}  \\ \hline
							\rowcolor{gray!40}
							\textbf{MVS}  & \textbf{Dep} & \textbf{Tra} & \textbf{Tran}  & \textbf{Test PSNR} $^\uparrow$ & \textbf{Test SSIM} $^\uparrow$ & \textbf{Test LPIPS} $^\downarrow$ \\ \hline 
							$\checkmark$     & $\times$ & $\times$  & $\times$               & 24.77 & 0.692 & 0.244 \\
							$\checkmark$     & $\checkmark$ & $\times$ & $\times$            & 25.07 & 0.752 & 0.241 \\  
							$\checkmark$     & $\checkmark$ & $\checkmark$ & $\times$        & 26.61 & 0.818 & 0.239 \\ 
							$\checkmark$     & $\checkmark$ & $\checkmark$ & $\checkmark$    & \textbf{26.67} & \textbf{0.819} & \textbf{0.235} \\ \hline
					\end{tabular}}
					\vspace{-0.6cm}
				\end{table}
				
				\begin{table}[t]
					\centering
					\footnotesize
					\vspace{-0.55cm}
					\caption{Ablation experiments of the effectiveness of different components in FingerNeRF on UNSW-3D dataset.}
					\label{tab:gen_nvs_ablation_unsw_3d_appendix}
					\setlength{\tabcolsep}{0.3mm}{
						\begin{tabular}{cccc|ccc}
							\hline 
							\rowcolor{gray!40}
							\textbf{MVS}  & \textbf{Dep} & \textbf{Tra} & \textbf{Tran}  & \textbf{Val PSNR} $^\uparrow$ & \textbf{Val SSIM} $^\uparrow$ & \textbf{Val LPIPS} $^\downarrow$  \\ \hline 
							$\checkmark$     & $\times$ & $\times$  & $\times$               & 27.57 & 0.499 & 0.213  \\
							$\checkmark$     & $\checkmark$ & $\times$ & $\times$            & 28.40 & 0.762 & 0.082  \\  
							$\checkmark$     & $\checkmark$ & $\checkmark$ & $\times$        & 29.09 & 0.738 & 0.077  \\ 
							$\checkmark$     & $\checkmark$ & $\checkmark$ & $\checkmark$    & \textbf{31.32} & \textbf{0.813} & \textbf{0.062}  \\ \hline
							\rowcolor{gray!40}
							\textbf{MVS}  & \textbf{Dep} & \textbf{Tra} & \textbf{Tran}  & \textbf{Test PSNR} $^\uparrow$ & \textbf{Test SSIM} $^\uparrow$ & \textbf{Test LPIPS} $^\downarrow$ \\ \hline 
							$\checkmark$     & $\times$ & $\times$  & $\times$               & 29.09 & 0.567 & 0.193 \\
							$\checkmark$     & $\checkmark$ & $\times$ & $\times$            & 28.49 & 0.763 & 0.088 \\  
							$\checkmark$     & $\checkmark$ & $\checkmark$ & $\times$        & 29.28 & 0.731 & 0.078 \\ 
							$\checkmark$     & $\checkmark$ & $\checkmark$ & $\checkmark$    & \textbf{30.97} & \textbf{0.801} & \textbf{0.064} \\ \hline
					\end{tabular}}
					% 	\vspace{-0.4cm}
				\end{table}
				
				\begin{table*}[t]
					\centering
					\caption{Comparison with Mulit-view 3D shape recognition methods on SCUT-Finger-3D dataset.}
					\label{tab:multi_view_comp_scut_finger_3d_appendix}
					\resizebox{0.9\linewidth}{!}{
						\begin{tabular}{l|ccccccc}
							\hline
							\rowcolor{gray!40}
							\textbf{Method}                       & \textbf{EER} $^\downarrow$       & \textbf{T-F=0.01} $^\uparrow$  & \textbf{T-F=0.001} $^\uparrow$ & \textbf{mAP} $^\uparrow$       & \textbf{Rank 1} $^\uparrow$    & \textbf{Rank 5} $^\uparrow$  \\ 
							\hline
							MVCNN \cite{su2015multi} + ResNet-18 \cite{he2016deep}            & 8.35\%    & 60.08\%   & 33.10\%   & 67.20\%   & 65.06\%   & 92.66\% \\
							MVCNN \cite{su2015multi} + ResNet-34 \cite{he2016deep}            & 8.97\%    & 60.42\%   & 42.38\%   & 66.85\%   & 63.33\%   & 90.35\% \\
							MVCNN \cite{su2015multi} + ResNet-50 \cite{he2016deep}            & 8.75\%    & 63.70\%   & 40.02\%   & 75.57\%   & 71.22\%   & 92.44\% \\
							MVCNN \cite{su2015multi} + ShuffleNet v2\_05 \cite{ma2018shufflenet}    & 14.21\%   & 41.70\%   & 30.98\%   & 61.71\%   & 53.69\%   & 80.74\% \\
							MVCNN \cite{su2015multi} + ShuffleNet v2\_10 \cite{ma2018shufflenet}   & 16.98\%   & 38.32\%   & 28.22\%   & 53.83\%   & 47.21\%   & 72.82\% \\
							MVCNN \cite{su2015multi} + MobileNet\_V2 \cite{sandler2018mobilenetv2}       & 5.08\%    & 80.92\%   & 44.70\%   & 84.07\%   & 78.72\%   & 95.96\% \\
							MVCNN \cite{su2015multi} + MobileNet\_V3\_large \cite{howard2019searching} & 6.01\%    & 72.68\%   & 49.46\%   & 80.75\%   & 77.69\%   & 94.55\% \\
							MVCNN \cite{su2015multi} + MobileNet\_V3\_small \cite{howard2019searching} & 7.44\%    & 69.54\%   & 42.00\%   & 76.96\%   & 73.94\%   & 92.37\% \\
							MVCNN \cite{su2015multi} + EfficientNet-B1 \cite{tan2019efficientnet}     & 5.44\%    & 77.30\%   & 51.32\%   & 80.00\%   & 80.06\%   & 95.87\% \\
							MVCNN \cite{su2015multi} + mnasnet\_05 \cite{tan2019mnasnet}         & 20.62\%   & 31.68\%   & 20.88\%   & 50.48\%   & 39.52\%   & 66.09\% \\
							MVCNN \cite{su2015multi} + mnasnet\_10 \cite{tan2019mnasnet}         & 12.20\%   & 50.46\%   & 34.29\%   & 64.30\%   & 62.15\%   & 89.39\% \\
							MVCNN \cite{su2015multi} + regnet\_y\_400mf \cite{radosavovic2020designing}     & 8.19\%    & 65.18\%   & 44.22\%   & 64.31\%   & 66.25\%   & 94.71\% \\
							MVCNN \cite{su2015multi} + regnet\_y\_800mf \cite{radosavovic2020designing}     & 6.75\%    & 70.18\%   & 42.82\%   & 74.34\%   & 71.96\%   & 92.20\% \\
							MVCNN \cite{su2015multi} + regnet\_x\_400mf \cite{radosavovic2020designing}     & 8.91\%    & 58.62\%   & 41.62\%   & 61.68\%   & 61.92\%   & 89.55\% \\
							MVCNN \cite{su2015multi} + regnet\_x\_800mf \cite{radosavovic2020designing}     & 7.66\%    & 67.60\%   & 39.12\%   & 72.00\%   & 68.72\%   & 93.75\% \\ 
							RotNet \cite{kanezaki2018rotationnet} + ResNet 18 \cite{he2016deep}           & 5.67\%    & 77.20\%   & 45.60\%   & 79.78\%   & 75.71\%   & 96.99\% \\
							RotNet \cite{kanezaki2018rotationnet} + ResNet 34 \cite{he2016deep}           & 7.32\%	 & 70.70\%	 & 45.86\%	 & 74.07\%   & 73.33\%   & 95.74\% \\
							RotNet \cite{kanezaki2018rotationnet} + ResNet 50 \cite{he2016deep}           & 5.65\% & 75.88\% & 41.40\% & 81.01\% & 76.89\% & 96.03\% \\
							RotNet \cite{kanezaki2018rotationnet} + ShuffleNet v2\_05 \cite{ma2018shufflenet}   & 14.50\%	 & 38.52\%	 & 22.66\%	 & 48.60\%	 & 47.02\%	 & 78.81\%   \\ 
							RotNet \cite{kanezaki2018rotationnet} + ShuffleNet v2\_10 \cite{ma2018shufflenet}   & 17.56\%	 & 34.08\%	 & 13.70\%	 & 42.41\%	 & 40.35\%	 & 70.80\%   \\
							RotNet \cite{kanezaki2018rotationnet} + MobileNet\_V2 \cite{sandler2018mobilenetv2}       & 5.00\%	 & 81.30\%	 & 56.94\%	 & 85.87\%	 & 82.15\%	 & 96.03\%   \\
							RotNet \cite{kanezaki2018rotationnet} + MobileNet\_V3\_large \cite{howard2019searching} & 5.69\%	 & 78.22\%	 & 55.82\%	 & 82.70\%	 & 80.06\%	 & 95.54\%   \\
							RotNet \cite{kanezaki2018rotationnet} + MobileNet\_V3\_small \cite{howard2019searching} & 5.80\%	 & 73.40\%	 & 43.62\%	 & 79.35\%	 & 74.07\%	 & 95.87\%   \\
							RotNet \cite{kanezaki2018rotationnet} + EfficientNet-B1 \cite{tan2019efficientnet}     & 5.07\%	 & 79.28\%	 & 55.26\%	 & 89.26\%	 & 84.39\%	 & 96.24\%   \\
							RotNet \cite{kanezaki2018rotationnet} + mnasnet\_05 \cite{tan2019mnasnet}         & 30.34\%	 & 31.50\%	 & 27.74\%	 & 37.30\%	 & 37.92\%	 & 55.61\%   \\
							RotNet \cite{kanezaki2018rotationnet} + mnasnet\_10 \cite{tan2019mnasnet}         & 26.81\%	 & 34.04\%	 & 28.00\%	 & 47.91\%	 & 38.94\%	 & 54.49\%   \\
							MHBN \cite{yu2018multi}                         & 28.25\%	 & 29.74\%	 & 23.68\%	 & 33.66\%	 & 34.65\%	 & 50.03\%   \\
							MHBN \cite{yu2018multi} + ResNet 18  \cite{he2016deep}            & 18.76\%	 & 30.04\%	 & 16.96\%	 & 38.65\%	 & 36.86\%	 & 64.68\%   \\
							MHBN \cite{yu2018multi} + ResNet 34  \cite{he2016deep}            & 21.52\%	 & 26.92\%	 & 12.42\%	 & 34.96\%	 & 34.10\%	 & 63.49\%   \\
							MHBN \cite{yu2018multi} + ResNet 50  \cite{he2016deep}            & 18.29\%	 & 33.08\%	 & 20.16\%	 & 37.37\%	 & 35.19\%	 & 66.28\%   \\
							MHBN \cite{yu2018multi} + ShuffleNet v2\_05 \cite{ma2018shufflenet}     & 18.38\%	 & 23.70\%	 & 12.22\%	 & 41.18\%	 & 37.15\%	 & 68.46\%   \\
							MHBN \cite{yu2018multi} + ShuffleNet v2\_10 \cite{ma2018shufflenet}     & 21.44\%	 & 20.58\%	 & 5.62\%	 & 38.94\%	 & 32.88\%	 & 65.03\%   \\
							MHBN \cite{yu2018multi} + MobileNet\_V2 \cite{sandler2018mobilenetv2}         & 20.01\%	 & 33.52\%	 & 13.14\%	 & 48.18\%	 & 38.65\%	 & 66.15\%   \\
							MHBN \cite{yu2018multi} + MobileNet\_V3\_large  \cite{howard2019searching} & 21.10\%	 & 27.06\%	 & 17.64\%	 & 50.71\%	 & 41.73\%	 & 62.69\%   \\
							MHBN \cite{yu2018multi} + MobileNet\_V3\_small  \cite{howard2019searching} & 21.06\%	 & 32.92\%	 & 24.96\%	 & 44.40\%	 & 39.20\%	 & 65.99\%   \\
							MHBN \cite{yu2018multi} + EfficientNet-B1
							\cite{tan2019efficientnet}      & 19.23\%	 & 31.30\%	 & 16.00\%	 & 40.05\%	 & 40.26\%	 & 65.26\%   \\
							\hline
							Yang. \cite{yang2021lfmb}                              & 8.35\%   & 60.08\%    & 33.10\%   & 67.20\%   & 65.06\%   & 92.66\% \\
							Lin. \cite{lin2018contactless} + ResNet 18 \cite{he2016deep}            & 5.61\%	& 77.10\%	 & 52.60\%	 & 78.49\%   & 75.93\%   & 96.47\% \\ 
							Lin. \cite{lin2018contactless} + ResNet 34 \cite{he2016deep}            & 6.19\%	& 73.80\%	 & 43.78\%	 & 78.51\%   & 74.74\%   & 96.44\% \\ 
							Lin. \cite{lin2018contactless} + ResNet 50 \cite{he2016deep}            & 6.14\%	& 69.44\%	 & 48.62\%	 & 75.25\%   & 73.17\%   & 95.87\% \\ \hline
							FingerNeRF (Ours) & \textbf{4.37\%} & \textbf{81.54\%} & \textbf{60.62\%} & \textbf{88.84\%} & \textbf{83.88\%} & \textbf{97.05\%} \\
							\hline
						\end{tabular}
					}
					\vspace{-0.4cm}
				\end{table*}
				
				\begin{table*}[t]
					\centering
					\caption{Comparison with Mulit-view 3D shape recognition methods on SCUT-FingerVein-3D dataset.}
					\label{tab:multi_view_comp_scut_fingervein_3d_appendix}
					\resizebox{0.9\linewidth}{!}{
						\begin{tabular}{l|ccccccc}
							\hline
							\rowcolor{gray!40}
							\textbf{Method}                       & \textbf{EER} $^\downarrow$       & \textbf{T-F=0.01} $^\uparrow$  & \textbf{T-F=0.001} $^\uparrow$ & \textbf{mAP} $^\uparrow$       & \textbf{Rank 1} $^\uparrow$    & \textbf{Rank 5} $^\uparrow$  \\ 
							\hline
							MVCNN \cite{su2015multi} + ResNet-18 \cite{he2016deep}            & 12.52\% & 48.25\% & 32.92\% & 50.73\% & 50.00\% & 81.22\% \\
							MVCNN \cite{su2015multi} + ResNet-34 \cite{he2016deep}            & 10.88\% & 49.17\% & 31.28\% & 62.09\% & 60.31\% & 86.48\% \\
							MVCNN \cite{su2015multi} + ResNet-50 \cite{he2016deep}            & 11.75\% & 43.30\% & 27.92\% & 57.30\% & 57.54\% & 85.30\% \\ 
							MVCNN \cite{su2015multi} + ShuffleNet v2\_05 \cite{ma2018shufflenet}    & 15.92\% & 36.13\% & 24.42\% & 42.05\% & 40.65\% & 68.54\% \\
							MVCNN \cite{su2015multi} + ShuffleNet v2\_10 \cite{ma2018shufflenet}   & 17.48\% & 29.09\% & 17.74\% & 38.14\% & 37.39\% & 66.03\% \\
							MVCNN \cite{su2015multi} + MobileNet\_V2 \cite{sandler2018mobilenetv2}      & 10.75\% & 55.45\% & 35.26\% & 61.59\% & 62.59\% & 88.23\% \\
							MVCNN \cite{su2015multi} + MobileNet\_V3\_large \cite{howard2019searching} & 12.63\% & 50.85\% & 31.92\% & 55.39\% & 55.87\% & 80.85\% \\
							MVCNN \cite{su2015multi} + MobileNet\_V3\_small \cite{howard2019searching} & 9.65\% & 50.85\% & 28.15\% & 60.15\% & 61.43\% & 89.93\% \\
							MVCNN \cite{su2015multi} + EfficientNet-B1 \cite{tan2019efficientnet}     &  9.06\% & 62.85\% & 28.65\% & 63.99\% & 63.63\% & 89.75\% \\
							MVCNN \cite{su2015multi} + mnasnet\_05 \cite{tan2019mnasnet}         & 15.49\% & 29.43\% &  0.00\% & 35.01\% & 28.22\% & 64.25\% \\
							MVCNN \cite{su2015multi} + mnasnet\_10 \cite{tan2019mnasnet}         & 12.42\% & 46.38\% & 27.04\% & 52.49\% & 54.33\% & 54.33\% \\
							MVCNN \cite{su2015multi} + regnet\_y\_400mf \cite{radosavovic2020designing}     & 10.78\% & 52.38\% & 31.75\% & 60.99\% & 57.99\% & 84.30\% \\
							MVCNN \cite{su2015multi} + regnet\_y\_800mf \cite{radosavovic2020designing}     &  9.24\% & 49.45\% & 29.19\% & 58.86\% & 55.93\% & 88.93\% \\
							MVCNN \cite{su2015multi} + regnet\_x\_400mf \cite{radosavovic2020designing}     & 11.50\% & 50.87\% & 26.85\% & 57.96\% & 53.54\% & 84.21\% \\
							MVCNN \cite{su2015multi} + regnet\_x\_800mf \cite{radosavovic2020designing}     & 11.27\% & 42.96\% & 27.17\% & 58.99\% & 52.54\% & 84.18\% \\ 
							RotNet \cite{kanezaki2018rotationnet} + ResNet 18 \cite{he2016deep}           & 10.62\% & 51.42\% & 33.72\% & 61.27\% & 61.27\% & 87.24\% \\
							RotNet \cite{kanezaki2018rotationnet} + ResNet 34 \cite{he2016deep}           & 10.62\% & 55.25\% & 30.00\% & 30.00\% & 55.60\% & 85.03\% \\
							RotNet \cite{kanezaki2018rotationnet} + ResNet 50 \cite{he2016deep}           & 10.42\% & 46.64\% & 28.21\% & 58.35\% & 53.81\% & 85.03\% \\
							RotNet \cite{kanezaki2018rotationnet} + ShuffleNet v2\_05 \cite{ma2018shufflenet}   & 15.22\% & 29.03\% & 17.76\% & 27.11\% & 24.83\% & 64.16\%   \\ 
							RotNet \cite{kanezaki2018rotationnet} + ShuffleNet v2\_10 \cite{ma2018shufflenet}   & 14.23\% & 21.64\% &  0.00\% & 32.37\% & 28.37\% & 70.30\%  \\
							RotNet \cite{kanezaki2018rotationnet} + MobileNet\_V2 \cite{sandler2018mobilenetv2}     & 9.23\% & 52.17\% & 32.30\% & 63.81\% & 63.81\% & 89.05\% \\
							RotNet \cite{kanezaki2018rotationnet} + MobileNet\_V3\_large \cite{howard2019searching} & 8.90\% & 59.49\% & 35.77\% & 62.86\% & 61.92\% & 91.32\% \\
							RotNet \cite{kanezaki2018rotationnet} + MobileNet\_V3\_small \cite{howard2019searching} & 9.52\% & 53.42\% & 33.40\% & 63.07\% & 63.07\% & 86.84\% \\
							RotNet \cite{kanezaki2018rotationnet} + EfficientNet-B1 \cite{tan2019efficientnet}     &  8.76\% & 60.98\% & 34.58\% & 66.63\% & 65.91\% & 89.96\%   \\
							RotNet \cite{kanezaki2018rotationnet} + mnasnet\_05 \cite{tan2019mnasnet}         & 34.48\% & 20.45\% & 13.74\% & 25.02\% & 21.02\% & 42.74\%   \\
							RotNet \cite{kanezaki2018rotationnet} + mnasnet\_10 \cite{tan2019mnasnet}         & 36.59\% & 19.65\% & 12.25\% & 23.01\% & 20.19\% & 39.74\%   \\
							MHBN \cite{yu2018multi}                         & 36.59\% & 19.65\% & 12.25\% & 23.01\% & 20.19\% & 39.74\%   \\
							MHBN \cite{yu2018multi} + ResNet 18  \cite{he2016deep}            & 20.16\% & 22.30\% &  7.32\% & 34.96\% & 32.73\% & 62.46\%  \\
							MHBN \cite{yu2018multi} + ResNet 34  \cite{he2016deep}            & 23.54\% & 20.13\% & 10.32\% & 26.91\% & 27.86\% & 53.42\%   \\
							MHBN \cite{yu2018multi} + ResNet 50  \cite{he2016deep}            & 18.86\% & 21.00\% & 15.02\% & 30.46\% & 31.76\% & 63.01\%   \\
							MHBN \cite{yu2018multi} + ShuffleNet v2\_05 \cite{ma2018shufflenet}     & 17.62\% & 25.00\% & 11.77\% & 32.10\% & 33.27\% & 60.89\%   \\
							MHBN \cite{yu2018multi} + ShuffleNet v2\_10 \cite{ma2018shufflenet}     & 18.77\% & 12.66\% &  3.91\% & 29.84\% & 28.68\% & 60.01\%   \\
							MHBN \cite{yu2018multi} + MobileNet\_V2 \cite{sandler2018mobilenetv2}      & 38.75\% & 4.04\% & 4.20\% & 4.20\% & 5.93\% & 22.08\%   \\
							MHBN \cite{yu2018multi} + MobileNet\_V3\_large  \cite{howard2019searching} & 34.00\% & 3.23\% & 4.10\% & 8.53\% & 9.17\% & 19.20\%   \\
							MHBN \cite{yu2018multi} + MobileNet\_V3\_small  \cite{howard2019searching} & 32.00\% & 9.08\% & 3.75\% & 6.40\% & 6.53\% & 30.22\%  \\
							MHBN \cite{yu2018multi} + EfficientNet-B1 \cite{tan2019efficientnet}      & 35.99\% &  6.13\% &  1.51\% &  8.77\% & 10.28\% & 29.70\%   \\
							\hline
							Yang. \cite{yang2021lfmb}                                         & 11.91\% & 44.13\% & 25.38\% & 46.57\% & 45.64\% & 83.39\% \\
							Lin. \cite{lin2018contactless} + ResNet 18 \cite{he2016deep}      & 10.38\% & 53.21\% & 34.30\% & 58.23\% & 56.11\% & 85.00\% \\ 
							Lin. \cite{lin2018contactless} + ResNet 34 \cite{he2016deep}      & 11.82\% & 49.75\% & 33.81\% & 62.14\% & 55.69\% & 80.79\% \\ 
							Lin. \cite{lin2018contactless} + ResNet 50 \cite{he2016deep}      & 11.28\% & 49.38\% & 32.38\% & 49.38\% & 55.44\% & 85.72\% \\ \hline
							FingerNeRF (Ours)                                                     & \textbf{8.12\%} & \textbf{63.19\%} & \textbf{36.68\%} & \textbf{68.35\%} & \textbf{66.00\%} & \textbf{91.89\%} \\
							\hline
						\end{tabular}
					}
					\vspace{-0.4cm}
				\end{table*}
				
				\begin{table*}[t]
					\centering
					\caption{Comparison with Mulit-view 3D shape recognition methods on UNSW-3D dataset.}
					\label{tab:multi_view_comp_unsw_3d_appendix}
					\resizebox{0.9\linewidth}{!}{
						\begin{tabular}{l|ccccccc}
							\hline
							\rowcolor{gray!40}
							\textbf{Method}                       & \textbf{EER} $^\downarrow$       & \textbf{T-F=0.01} $^\uparrow$  & \textbf{T-F=0.001} $^\uparrow$ & \textbf{mAP} $^\uparrow$       & \textbf{Rank 1} $^\uparrow$    & \textbf{Rank 5} $^\uparrow$  \\ 
							\hline
							MVCNN \cite{su2015multi} + ResNet-18 \cite{he2016deep}            & 17.48\%	& 53.67\%	& 32.52\%	& 15.64\%	& 14.65\%	& 25.83\% \\
							MVCNN \cite{su2015multi} + ResNet-34 \cite{he2016deep}            & 21.49\%	& 43.21\%	& 26.06\%	& 13.71\%	& 13.98\%	& 22.77\% \\
							MVCNN \cite{su2015multi} + ResNet-50 \cite{he2016deep}            & 16.82\%	& 55.68\%	& 39.87\%	& 16.13\%	& 15.98\%	& 25.97\% \\
							MVCNN \cite{su2015multi} + ShuffleNet v2\_05 \cite{ma2018shufflenet}    & 11.14\%	& 66.82\%	& 55.46\%	& 18.14\%	& 21.17\%	& 35.02\% \\
							MVCNN \cite{su2015multi} + ShuffleNet v2\_10 \cite{ma2018shufflenet}   & 13.70\%	& 66.15\%	& 48.78\%	& 18.98\%	& 17.98\%	& 30.36\% \\
							MVCNN \cite{su2015multi} + MobileNet\_V2 \cite{sandler2018mobilenetv2}       & 3.56\%	& 92.65\%	& 81.96\%	& 24.37\%	& 24.37\%	& 37.28\% \\
							MVCNN \cite{su2015multi} + MobileNet\_V3\_large \cite{howard2019searching} & 7.57\%	& 78.17\%	& 43.88\%	& 20.77\%	& 19.84\%	& 33.29\% \\
							MVCNN \cite{su2015multi} + MobileNet\_V3\_small \cite{howard2019searching} & 5.35\%	& 88.86\%	& 78.62\%	& 22.90\%	& 24.37\%	& 39.15\% \\
							MVCNN \cite{su2015multi} + EfficientNet-B1 \cite{tan2019efficientnet}     & 5.57\%	& 86.41\%	& 76.17\%	& 22.33\%	& 24.50\%	& 38.75\% \\
							MVCNN \cite{su2015multi} + mnasnet\_05 \cite{tan2019mnasnet}         & 8.69\%	& 77.06\%	& 61.92\%	& 20.46\%	& 22.10\%	& 32.36\% \\
							MVCNN \cite{su2015multi} + mnasnet\_10 \cite{tan2019mnasnet}         & 8.57\%	& 83.07\%	& 69.27\%	& 21.21\%	& 23.97\%	& 35.29\% \\
							MVCNN \cite{su2015multi} + regnet\_y\_400mf \cite{radosavovic2020designing}     & 13.36\%	& 67.48\%	& 20.04\%	& 15.01\%	& 17.44\%	& 29.69\% \\
							MVCNN \cite{su2015multi} + regnet\_y\_800mf \cite{radosavovic2020designing}     & 9.13\%	& 67.71\%	& 18.71\%	& 13.66\%	& 15.71\%	& 28.50\% \\
							MVCNN \cite{su2015multi} + regnet\_x\_400mf \cite{radosavovic2020designing}     & 11.25\%	& 71.49\%	& 52.78\%	& 17.12\%	& 18.38\%	& 28.89\% \\
							MVCNN \cite{su2015multi} + regnet\_x\_800mf \cite{radosavovic2020designing}     & 12.58\%	& 71.49\%	& 58.80\%	& 16.16\%	& 18.38\%	& 30.36\% \\ 
							RotNet \cite{kanezaki2018rotationnet} + ResNet 18 \cite{he2016deep}           & 4.01\%	& 89.53\%	& 79.51\%	& 22.84\%	& 23.57\%	& 35.95\% \\
							RotNet \cite{kanezaki2018rotationnet} + ResNet 34 \cite{he2016deep}           & 3.67\%	& 91.98\%	& 87.75\%	& 22.31\%	& 22.50\%	& 31.96\% \\
							RotNet \cite{kanezaki2018rotationnet} + ResNet 50 \cite{he2016deep}           & 3.56\%	& 93.32\%	& 81.74\%	& 21.92\%	& 24.37\%	& 37.42\% \\
							RotNet \cite{kanezaki2018rotationnet} + ShuffleNet v2\_05 \cite{ma2018shufflenet}   & 9.80\%	& 68.82\%	& 42.98\%	& 15.94\%	& 18.51\%	& 30.49\%  \\ 
							RotNet \cite{kanezaki2018rotationnet} + ShuffleNet v2\_10 \cite{ma2018shufflenet}   & 12.69\%	& 69.71\%	& 62.36\%	& 17.56\%	& 19.97\%	& 32.22\% \\
							RotNet \cite{kanezaki2018rotationnet} + MobileNet\_V2 \cite{sandler2018mobilenetv2}       & 3.56\%	& 92.65\%	& 81.96\%	& 24.37\%	& 24.37\%	& 37.28\% \\
							RotNet \cite{kanezaki2018rotationnet} + MobileNet\_V3\_large \cite{howard2019searching} & 4.23\%	& 92.65\%	& 81.74\%	& 23.20\%	& 23.70\%	& 35.55\%  \\
							RotNet \cite{kanezaki2018rotationnet} + MobileNet\_V3\_small \cite{howard2019searching} & 5.68\%	& 88.42\%	& 83.52\%	& 18.81\%	& 21.30\%	& 31.56\%  \\
							RotNet \cite{kanezaki2018rotationnet} + EfficientNet-B1 \cite{tan2019efficientnet}     & 3.79\% & 94.21\% & 78.62\% & 23.89\% & 24.50\% & 39.41\% \\
							RotNet \cite{kanezaki2018rotationnet} + mnasnet\_05 \cite{tan2019mnasnet}         & 8.57\%	& 78.84\%	& 60.58\%	& 20.80\%	& 21.70\%	& 31.16\%   \\
							RotNet \cite{kanezaki2018rotationnet} + mnasnet\_10 \cite{tan2019mnasnet}         & 8.69\%	& 85.52\%	& 73.72\%	& 20.45\%	& 21.44\%	& 33.95\%   \\
							MHBN \cite{yu2018multi}                        & 27.39\%	& 10.02\%	& 0.45\%	& 5.85\%	& 4.79\%	& 10.92\%  \\
							MHBN \cite{yu2018multi} + ResNet 18  \cite{he2016deep}            & 24.94\%	& 34.52\%	& 25.84\%	& 5.91\%	& 5.86\%	& 14.38\%  \\
							MHBN \cite{yu2018multi} + ResNet 34  \cite{he2016deep}            & 20.38\%	& 40.76\%	& 32.07\%	& 6.72\%	& 7.32\%	& 15.71\%  \\
							MHBN \cite{yu2018multi} + ResNet 50  \cite{he2016deep}            & 29.96\%	& 16.70\%	& 1.11\%	& 5.18\%	& 5.19\%	& 12.25\%  \\
							MHBN \cite{yu2018multi} + ShuffleNet v2\_05 \cite{ma2018shufflenet}     & 22.83\%	& 24.72\%	& 11.14\%	& 6.84\%	& 5.73\%	& 16.11\% \\
							MHBN \cite{yu2018multi} + ShuffleNet v2\_10 \cite{ma2018shufflenet}     & 21.60\%	& 27.39\%	& 1.34\%	& 5.34\%	& 4.93\%	& 14.38\% \\
							MHBN \cite{yu2018multi} + MobileNet\_V2 \cite{sandler2018mobilenetv2}         & 18.37\%	& 42.09\%	& 16.48\%	& 8.85\%	& 8.92\%	& 18.51\%   \\
							MHBN \cite{yu2018multi} + MobileNet\_V3\_large  \cite{howard2019searching} &  16.48\%	& 47.88\%	& 26.28\%	& 8.50\%	& 11.05\%	& 22.10\%  \\
							MHBN \cite{yu2018multi} + MobileNet\_V3\_small  \cite{howard2019searching} & 19.82\%	& 38.31\%	& 32.52\%	& 8.18\%	& 9.59\%	& 19.71\%  \\
							MHBN \cite{yu2018multi} + EfficientNet-B1 \cite{tan2019efficientnet}      & 21.94\%	& 31.40\%	& 5.79\%	& 8.27\%	& 7.32\%	& 15.98\%  \\
							\hline
							Yang. \cite{yang2021lfmb}                              & 16.37\%	& 41.20\%	& 26.95\%	& 12.47\%	& 10.65\%	& 19.84\% \\
							Lin. \cite{lin2018contactless} + ResNet 18 \cite{he2016deep}            & 5.46\%	& 85.52\%	& 60.13\%	& 18.74\%	& 20.77\%	& 31.42\% \\ 
							Lin. \cite{lin2018contactless} + ResNet 34 \cite{he2016deep}            & 7.35\%	& 71.94\%	& 58.35\%	& 13.89\%	& 16.11\%	& 25.97\% \\ 
							Lin. \cite{lin2018contactless} + ResNet 50 \cite{he2016deep}            & 4.01\%	& 92.43\%	& 86.19\%	& 23.77\%	& 23.44\%	& 34.89\% \\ \hline
							FingerNeRF (Ours) & \textbf{2.90\%} & \textbf{94.65\%} & \textbf{87.31\%} & \textbf{25.49\%} & \textbf{25.30\%} & \textbf{39.55\%} \\
							\hline
						\end{tabular}
					} 
					\vspace{-0.4cm}
				\end{table*}
\subsubsection{Comparison on SCUT-Finger-3D}
			\label{experiment:multi_view:scut_finger_3d}
			
			In this section, we evaluate the multi-view recognition methods on SCUT-Finger-3D.
			The qualitative results are provided in Table \ref{tab:multi_view_comp_scut_finger_3d}.
			The verification metrics includes EER, T-F=0.01, and T-F=0.001.
			The identification metrics includes mAP, Rank 1, and Rank 5.
			From the comparison with various multi-view recognition methods and the customized multi-view methods for finger traits, we can find that our proposed method can significantly outperform these methods in all metrics.
			Note that our FingerNeRF is supervised by differentiable neural rendering through original images, that does not require any 3D ground truth or customized device.
			Our method can involve the geometric information effectively, achieving 4.37\% EER and 88.84\% mAP.
			The experimental results prove that the proposed method can be simply combined with multi-view recognition network, and achieve superior performance by involving extra geometric information through the self-supervised pretrained FingerNeRF.

			\subsubsection{Comparison on SCUT-FingerVein-3D}
			\label{experiment:multi_view:scut_fingervein_3d}
			
			In this section, we migrate the experiments in previous section from the raw finger image to the finger vein images.
			As shown in Table \ref{tab:multi_view_comp_scut_fingervein_3d}, the qualitative comparison among multi-view recognition methods and the proposed method are provided.
			In ananology with Section \ref{experiment:multi_view:scut_finger_3d}, the verification metric of EER/T-F=0.01/T-F=0.001 and the identification metric of mAP/Rank 1/Rank 5 are used.
			From the table, we can find that on all 6 evaluation metrics the proposed method achieve better performance compared with other multi-view recognition methods.
			The EER and mAP of our method are respectively 8.12\% and 68.35\%.
			These experiment proves that out proposed FingerNeRF can also be generalized to a different modality of finger vein images and achieves superior performance by involving extra 3D geometric features compared with other methods.
        
         \subsubsection{Comparison on UNSW-3D}
	\label{experiment:multi_view:unsw_3d}
		In this section, we evaluate the verification and identification performance when generalized to existing multi-view finger trait dataset, UNSW-3D.
			In Table \ref{tab:multi_view_comp_unsw_3d}, the qualitative results are provided for comparison.
			As the results show, our method achieves better performance compared with other exisiting methods on all metrics of EER, T-F=0.01, T-F=0.001, mAP, Rank 1, and Rank 5.
			It proves that our proposed method can also be generalized to existing multi-view finger trait dataset and boost the performance with the learnable implicit 3D representation through self-supervision by neural rendering.
		\subsection{Generalizable Neural Rendering}	
				
				In the main manuscript (Section \ref{experiment:generalizable_rendering}), we provide the quantitative results of generalizable neural rendering on the test set of 3 different datasets: SCUT-Finger-3D, SCUT-FingerVein-3D, and UNSW-3D.
				Here, we further provide the quantitative results on the validation set for a complete evaluation.
				The complete experimental results are provided in the following tables (Table \ref{tab:gen_nvs_scut_finger_3d_appendix}, \ref{tab:gen_nvs_scut_fingervein_3d_appendix}, \ref{tab:gen_nvs_unsw_3d_appendix}.)
				The quantitative results on SCUT-Finger-3D, SCUT-FingerVein-3D, and UNSW-3D are respectively provided in Table \ref{tab:gen_nvs_scut_finger_3d_appendix}, \ref{tab:gen_nvs_scut_fingervein_3d_appendix}, \ref{tab:gen_nvs_unsw_3d_appendix}.
				From the tables, we can witness significant improvement on the validation PSNR, SSIM, and LPIPS on all datasets.
				This finding is similar as the discussion of the test PSNR, SSIM, and LPIPS on the test set in the main manuscript.

				\subsection{Ablation Study on Generalizable Neural Rendering}
				
				In the main manuscript (Section \ref{experiment:generalizable_rendering:ablation}), we provide the ablation results on the test set of the utilized datasets to evaluate the performance on generalizable neural rendering.
				Here, we further provide the ablation results on the validation set of the three utilized datasets: SCUT-Finger-3D, SCUT-FingerVein-3D, and UNSW-3D.
				As shown in Table \ref{tab:gen_nvs_ablation_scut_finger_3d_appendix}, \ref{tab:gen_nvs_ablation_scut_fingervein_3d_appendix}, \ref{tab:gen_nvs_ablation_unsw_3d_appendix}, we respectively present the ablation experiments of the effectiveness of different components in our FingerNeRF on each of the aforementioned datasets.
				From the tables, we can find that each components can improve the rendering performance effectively on the validation set as well as the test set.

				\subsection{Multi-view Recognition}
				In the main manuscript (Section \ref{experiment:multi_view}), we provide the comparison with other state-of-the-art multi-view-based recognition methods under the protocol introduced in Section \ref{experiment:multi_view:settings}.
				In Table \ref{tab:multi_view_comp_scut_finger_3d_appendix}, \ref{tab:multi_view_comp_scut_fingervein_3d_appendix}, \ref{tab:multi_view_comp_unsw_3d_appendix}, we respectively show the comparison results on SCUT-Finger-3D, SCUT-FingerVein-3D, and UNSW-3D datasets.
				In the main manuscript, we select some baseline to save the page space, while we provide a more complete comparison here.
				As the tables reflects, our method can effectively improve the multi-view recognition results in a generalizable setting.
				
			\section{Limitation}
			\label{limitation}
			Despite our proposed FingerNeRF can achieve great improvement compared with explicit 3D recognition methods, it still requires nonnegligible efforts in: 1) data pre-processing before training and 2) relatively long time for training.
			First, it requires elaborate data-preprocessing operations. 
			The finger trait data is different from the natural images in the wild, we have to remove the backgrounds from the supervision during training. 
			If the backgrounds are involved in the training process, the convergence of the network might be disturbed and fall into a trival solution.
			Second, the training cost might take several days due to the huge amount of unlabeled data.
			We naively utilize all unlabeled data during the training process without any data cleaning or curriculum strategies.
			Further filtering of the raw data and accelerating techniques might boost the training speed in the future.
			The non-uniform light condition may lead to poor reconstruction quality of the proposed FingerNeRF.
			The robustness towards non-uniform light conditions is an important problem of the future studies of our FingerNeRF.

			\section{Conclusion}
			\label{conclusion}
			
			In this paper, we propose a novel implicit representation for 3D finger biometrics.
			The demerits of existing explicit 3D methods in finger biometrics includes: 1) Information loss in 3D reconstruction; 2) Tight coupling between hardware and software. 
			Instead of following such an explicit reconstruct-and-recognize pipeline, we consider the problem in an implicit way.
			With the help of neural radiance fields (NeRF), the 3D representation can be implicitly handled by the neural network.
			However, the shape-radiance ambiguity problem may lead to incorrect 3D geometry, thus degrading the final performance.
			Consequently, we propose FingerNeRF, a novel generalizable NeRF specifically designed for 3D finger biometrics.
			First, we propose a novel Trait Guided Transformer to enhance the cross-view correspondence in the cost volume with the guidance of trait priors.
			Second, we involve extra geometric constraints into the neural rendering process via Depth Distillation Loss and Trait Guided Rendering Loss.
			Furthermore, we propose two new multi-view finger trait datasets with different modalitis: SCUT-Finger-3D and SCUT-FingerVein-3D.
			We conduct experiments on SCUT-Finger-3D, SCUT-FingerVein-3D, and UNSW-3D to evaluate the proposed method.
			The results prove that our proposed method can achieve superior performance.
			
		      \bibstyle{unsrt}
			\bibliography{sn-bibliography}% common bib file
			
			\clearpage

		\end{document}